\documentclass[alpha-refs]{wiley-article}

\usepackage{color}
\usepackage{ulem}

\usepackage{siunitx}

\usepackage{bm}
\usepackage{cancel}
\usepackage{algorithm}
\usepackage{algpseudocode}
\usepackage{subcaption}
\usepackage{hyperref}       
\usepackage{url}

\usepackage{bigints}

\papertype{Original Article}
\paperfield{Journal Section}

\title{Modelling calibration uncertainty in networks of environmental sensors}

\author[1]{Michael Thomas Smith}
\author[2]{Magnus Ross}
\author[3]{Joel Ssematimba}
\author[1]{Pablo A. Alvarado}
\author[2]{Mauricio \'{A}lvarez}
\author[3]{Engineer Bainomugisha}
\author[4]{Richard Wilkinson}

\affil[1]{Department of Computer Science, University of Sheffield, UK}
\affil[2]{Department of Computer Science, University of Manchester, UK}
\affil[3]{Department of Computer Science, Makerere University, Kampala, Uganda}
\affil[4]{School of Mathematical Sciences, University of Nottingham, UK}

\corraddress{Michael Thomas Smith, Department of Computer Science, University of Sheffield, Sheffield, S1 4DP, UK}
\corremail{m.t.smith@sheffield.ac.uk}

\fundinginfo{google.org.\\ EPSRC GCRF grant, EP/T00343X/1.} 

\runningauthor{Smith et al.}

\begin{document}

\maketitle

\begin{abstract}
Networks of low-cost sensors are becoming ubiquitous, but often suffer from poor accuracies and drift. Regular colocation with reference sensors allows recalibration but is complicated and expensive. Alternatively the calibration can be transferred using low-cost, mobile sensors. However inferring the calibration (with uncertainty) becomes difficult. We propose a variational approach to model the calibration across the network. We demonstrate the approach on synthetic and real air pollution data, and find it can perform better than the state of the art (multi-hop calibration). We extend it to categorical data produced by citizen-scientist labelling. In Summary: The method achieves uncertainty-quantified calibration, which has been one of the barriers to low-cost sensor deployment and citizen-science research.

\end{abstract}

\section{Introduction}
Large networks of low-cost sensors are becoming increasingly common across a range of domains, including weather \citep{van2014trans}, snow-cover \citep{pohl2014potential}, wildlife \citep{dyo2010evolution} and air pollution \citep{khedo2010wireless,liu2020low}. The aim is to collect far more data at low-cost. Unfortunately, across all these domains is the common issue of how to ensure low-cost sensors remain calibrated, while still achieving substantial cost-savings. Citizen science data collection initiatives often have similar issues, necessitating validation by experts, replication and statistical modelling \citep[see][for a review]{kosmala2016assessing}. In particular, the problem of maintaining calibration over time, in low cost sensor networks, typically requires regular colocation recalibration \citep{rai2017end}.


Considerable work has already been invested in developing approaches for the reliable use of low-cost sensors for air pollution monitoring (see Section \ref{background}). \cite{rai2017end} emphasise that the key issue with the use of low-cost sensors is data quality: For the network to have some use for policy makers the sensors need frequent calibration in `\emph{final deployment conditions}' such that sensor calibration drift should be accounted for.

This work is motivated by the need to manage the calibration of a network of low-cost PM2.5, air pollution sensors deployed in Kampala, Uganda. Kampala is a low-income, East African city, associated with a tropical, high-humidity, dusty environment; leading to additional issues with sensor reliability and drift. Calibration in this network is achieved through regular but brief visits to the static (low-cost and reference) sensors, by mobile low-cost sensors mounted on motorbike taxis. The motorbike taxi visits are often opportunistic and usually last less than ninety minutes. This leads to a complex network of in-situ colocation events. Our experience is that drift (due to dust, etc) experienced in Kampala is much greater than similar sensors running in the UK, suggesting that this should be included in any model. We found that most of the research on pollution monitoring were for cities in temperate climates. Very few environmental monitoring networks are on the continent of Africa \citep{mao2019low}. Hopefully the approach described will help guide future deployments across the continent.

\subsection{The task}

We have a set of particulate (PM2.5) sensors. Some are mobile and some are static. Some of the static sensors are `reference' instruments which we assume measure the true pollution. The rest are low-cost sensors. We assume that the true pollution is a parametric function, $\phi(\text{raw measurement}, \text{parameters})$, of the raw observation from the low-cost sensor. \emph{The task is to estimate this calibration function (or rather its vector of parameters) for each sensor, for any given time, to allow us to use the low-cost sensors to estimate the true air pollution in locations where there is no reference instrument.}

The full problem consists of both estimating the calibration functions \emph{and} modelling the pollution over space and time across the city. Modelling these two aspects jointly may have advantages, but we focus in this paper on simply solving the calibration problem, where each sensor-pair colocation occurs at an arbitrary, undefined spatial location.

For a given sensor, $s$ at time, $t$, we wish to find the value of the calibration function's parameter vector, $\bm{f}_{s}(t)$. We will estimate this using a list of colocations. Each colocation records: (a) The time of the colocation (b) The id of the two sensors (c) The raw (uncalibrated) measurements of the two sensors. To illustrate with an example, the calibration function could be a simple linear regression with two parameters (an offset and a gradient), such that: $\text{true pollution} = \text{raw measurement} \times \text{gradient} + \text{offset}$. The gradient and offset make up a vector of two parameters, which we wish to estimate. These we believe can change over time, and differ between sensors.
We want a method that: (1) Can infer the calibration function's parameters even if the associated sensor has not been directly visited by a reference sensor. (2) Can combine calibration information arriving through multiple pathways of calibration. (3) Uses both static and mobile sensor data. (4) Has a reasonable model of how the measurement is related to the true pollution. (5) Provides principled estimates of the uncertainty in the calibration, taking into account time.
%
\subsection{Background}
\label{background}
A review by \cite{delaine2019situ} divides the calibration literature in several ways, considering reference instrumentation, sensor mobility, types of calibration (time dependency and complexity) and `grouping strategies'. Calibration can occur prior to deployment or can occur in-situ by finding those times in which pollution is believed to be similar across large distances, for example when few local sources of pollution exist at night \citep[e.g.][]{miskell2018solution}. We are interested in those approaches which use physical post-deployment colocation.

Consider first \cite{arfire2015model}. They have a model-based approach which assumes a low-cost chemical sensor is calibrated by co-location with a reference sensor regularly. This doesn't then handle the more complex network that we are analysing (in which colocations between low-cost sensors also occurs). They consider temporal drift by either a linear or a hyperbolic term with respect to time. There is no indication that they compute or assess the uncertainty in the model's predictions.

Of more interest to our application are studies looking at networks of sensors such that the low-cost sensors are not necessarily directly calibrated by the reference instrument but potentially by intermediate sensors (e.g. mobile sensors). \cite{tsujita2004dynamic} considers one such model. However they make a series of assumptions (e.g. how the sensor calibration drift is a fixed linear number). The calibration is also done in a non-probabilistic approach; with an estimate for the pollution at a location simply the average of all sensors, even though some will, presumably, be more reliable than others (e.g. if they have just visited the reference station). The model also doesn't report its confidence. Setting aside these issues, the paper does suggest that a network approach to calibration might be of benefit. \cite{xiang2012collaborative} use the uncertainty in a given sensor's calibration when combining measurements. The problem this paper considers largely matches ours. One major difference is the error model. In our experience the calibration we are concerned with involves scaling of the true pollution, not just an offset. One potentially could alter their paper by using the log of sensor measurements, which will then make their sum a product. They also require that the calibration computation and data remains local to the device, while we are computing this using all the data from all sensors. \cite{markert2018privacy} assign weights to various estimates but, as with almost all the papers, the correlations between the calibration estimates are not considered. Their treatment is partly Bayesian (for example in how different uncertainties are combined), but has a single value for drift (subtracting the calibrated reading from the raw reading). Also by only looking at previous colocations, it fails to propagate all the calibration information available. We suggest therefore that there are likely to be considerable improvements in calibration accuracy and uncertainty quantification by using a process model of calibration over time, which will be able to leverage prior knowledge around temporal structure in calibration functions. A fully Bayesian treatment will also be able to handle more complex measurement functions.
%
%
These criticisms aside, \cite{markert2018privacy} provide a useful foundation and demonstration of the benefits of a network of sensors. Another related work \citep{hasenfratz2012fly} looks at a similar problem, this time using a polynomial fit for the sensor data (rather than just adding an offset). They handle multiple sensors by computing weighted sums, with weights dependent on the time difference (between query and calibration) although not in a particularly probabilistic manner. The paper computes a single ML estimate for the calibration.

Most other papers \citep[e.g.][]{maag2017scan,kizel2018node,bychkovskiy2003collaborative,fonollosa2016calibration} do not consider drift over time. Many, \citep[e.g.][]{maag2017scan,bychkovskiy2003collaborative} do not handle uncertainty in their calibration estimates. Finally note that most of the papers mentioned in this section assume linear/scaling-only calibration. In summary, there is currently no framework for performing calibration across an arbitrary network of sensor-rendezvous events which probabilistically handles the quantification of uncertainty in the calibration with temporal drift. The method described in this paper correctly estimates the uncertainty in calibration across a complex network consisting of mobile, static, low-cost and reference instruments.

\subsubsection{Related work in probabilistic modelling}

We did originally explore a joint model, in which both the calibration parameters \emph{and the pollution} were modelled as Gaussian processes. This had potentially more power as non-colocated (but still correlated) measurements could still inform the estimates of the calibration, but it was found to be somewhat intractable, and hence we focus in this paper on the calibration pair model  (in which we treat the colocations independently of their spatial locations) and leave the joint model for future work.

It is however worth considering the necessary tools for inference in the joint model, as it connects with other literature. The joint model involved a calibration function parameterised by a Gaussian process (modelling, e.g. the calibration offset) operating on another Gaussian process (the true pollution over time and space). This is reminiscent, if the calibration function was simply scaling the pollution, of \cite{wilson2012gaussian}, in which a weighted sum of GPs is computed, with the weights consisting of GPs themselves. They experimented with both MCMC (using elliptical slice sampling) and variational EM to perform the inference on the latent variables. We experimented with using MCMC for inference in both models (see supplementary) but found mixing was challenging as the network grew larger. They noted that even with their relatively simple model, Gibbs sampling would `mix poorly because of the tight correlations between the weights and the nodes'. We found that even with whitening, HMC and other tricks, the strong correlations in the posterior made it very difficult to perform inference. \cite{alvarado2017efficient} more recently computed a similarly structured product of Gaussian processes, and used Gauss-Hermite quadrature. In our experiments we used simple Monte Carlo to compute the expectations, following the approaches in \cite{hensman2013gaussian} and \cite{salimbeni2017doubly}.

\subsection{Calibration pair model}
\label{calpair}
\begin{figure}[t!]
  \centering    
    \includegraphics[width=0.78\columnwidth]{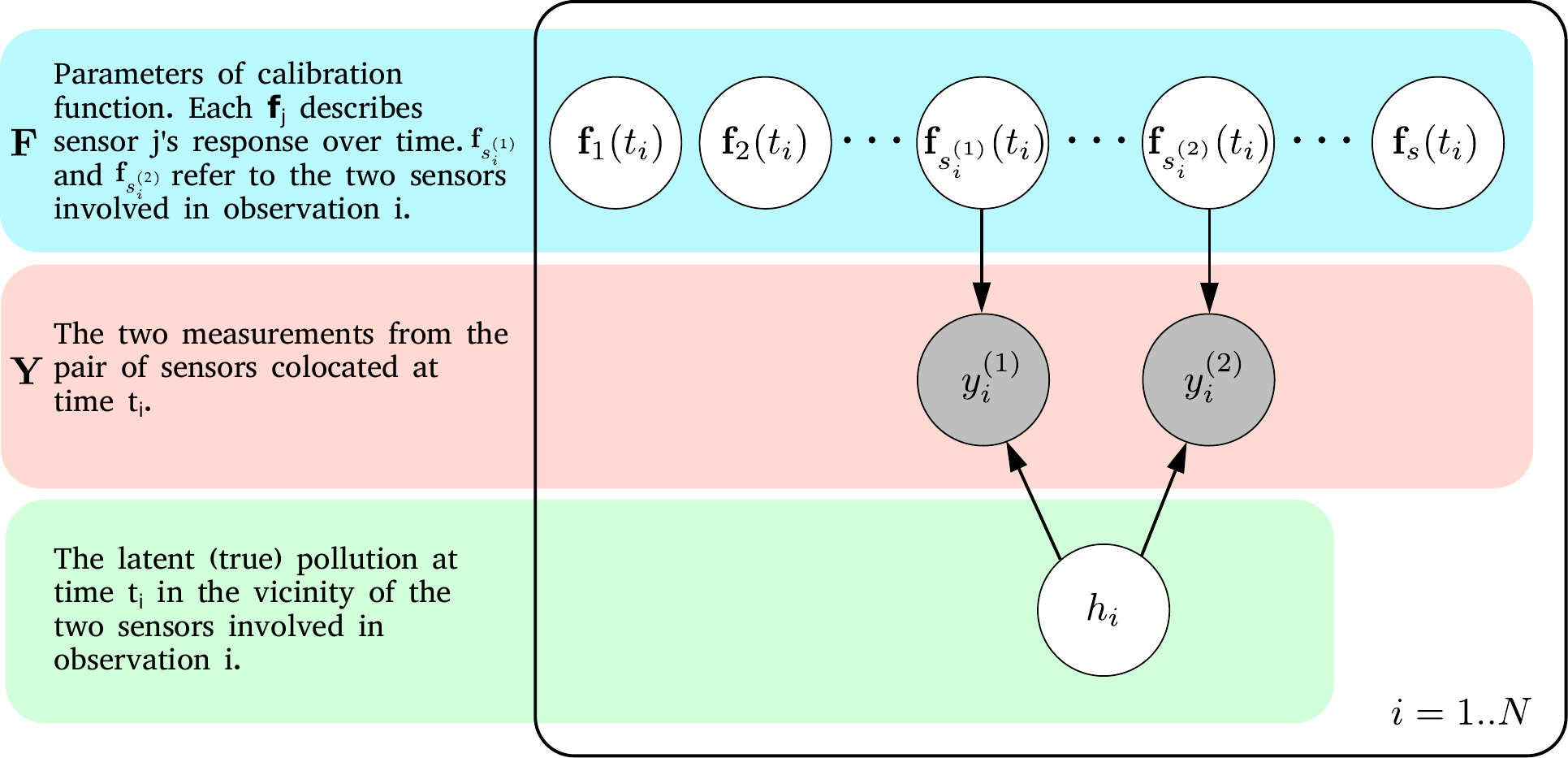}
  \caption{Calibration pair model. \textit{A priori} the vectors of GP random variables in $\bm{F}$ are independent, but become correlated in the posterior due to colocated observations connecting them together in pairs.}
  \label{plate}
\end{figure}

\subsubsection{Model definition}
Each of the $i=1..N$ observations, $\bm{Y} \in \mathcal{R}^{N \times 2}$, consist of two measurements, $y_i^{(1)}, y_i^{(2)}$, by two sensors with indices $s_i^{(1)},s_i^{(2)}$, colocated at time $t_i$. One could, in principle, have more than two sensors colocated, but for simplicity in representation and likelihood function we have used pair-wise colocation. The model assumes that a deterministic parametric function, $\phi$, describes how a measurement is related to the true pollution: 
$\text{true pollution} = \phi(\text{raw measurement}, \text{parameters}) + \text{unstructured noise}$.

For each sensor, $j$, the calibration function's parameter vector consists of $C$ parameters that are each functions of time: $\bm{f}_{j}(t)$. Our model assigns an independent Gaussian process prior (over time) to each of these $c=1..C$ parameters for each of the $j=1..S$ sensors. So, let $\bm{F}(t)$ be an $S \times C$ matrix of \emph{independent} Gaussian processes $[f_{j}]_c(t) \sim GP(0,k_{j,c})$: We are assuming, \textit{a priori}, that the calibration parameters of the sensors are independent between sensors and the parameters are also independent within sensors. In our implementation we have different kernels for sensors of different types and for different parameters, allowing us to model, for example, that one sensor type's offset drifts at a slower rate than its scaling, etc. The aim of the method is to infer the values of $\bm{F}(t)$.

We emphasise that, in this model we are discarding location information and just using information about which sensor pairs are colocated, and when.
\newcommand{\truepol}{h_i}

\subsubsection{The likelihood}
Here, by likelihood, we mean the probability of the observations, $\bm{Y}$, given the model, its hyper-parameters and any latent variables, $\bm{F}$.
To compute $\log p(\bm{Y}|\bm{F})$ we first assume that this can be factorised across observations: $p(\bm{Y}|\bm{F}) = \prod_{i=1}^N p(\bm{y}_i | \bm{F})$. Where $p(\bm{y}_i | \bm{F})$ is the probability of one pair of colocated observations, given our model and latent calibration parameters. 
%
%
We pick from $\bm{F}$ the two rows associated with the two sensors $s_i^{(1)}$ and $s_i^{(2)}$ that took the two measurements. So the calibrated predicted pollution from the two observations are, 
\newcommand{\f}[1]{\bm{f}_{s_i^{(#1)}}(t_i)}
\newcommand{\ft}[1]{\bm{f}_{s_i^{(#1)}}(t_i^{(#1)})}
\newcommand{\calfn}[1]{\phi \big(y_i^{(#1)},\f{#1} \big)}
$\calfn{1}$ and $\calfn{2}$.

Let $\truepol$ be the (latent) true pollution associated with the colocated observation pair $\bm{y}_i$. We assume that these calibration-corrected predictions of the pollution are normally distributed around the true (latent) pollution, i.e.: 
\begin{equation}
p\Bigg(\begin{bmatrix}\calfn{1}\\ \calfn{2}\end{bmatrix} \Bigg| h_i \Bigg) \sim N\Bigg(\begin{bmatrix}\truepol \\ \truepol\end{bmatrix}, \begin{bmatrix}\sigma^2 & 0 \\ 0 & \sigma^2 \end{bmatrix}\Bigg).
\end{equation}
If we place a wide, fairly uninformative, normal prior on $\truepol$ (with mean zero, and variance $\gamma^2$), we can integrate out $\truepol$ in closed form. Allowing us to write the probability of a given (calibrated) observation pair as a 2d joint multivariate Gaussian,
\begin{align}
p\Bigg(\begin{matrix}
           \calfn{1} \\
           \calfn{2} \\
         \end{matrix} \Bigg) & = \bigintss p\Bigg(\begin{bmatrix}\calfn{1}\\ \calfn{2}\end{bmatrix} \Bigg| h_i \Bigg)
         \times N\Big(\truepol \Big| 0, \gamma^2\Big) \;\;\;\text{d}\truepol\\
& \sim N\Bigg(0,\begin{bmatrix}
           \sigma^2 + \gamma^2 & \gamma^2 \\
           \gamma^2 & \sigma^2 + \gamma^2 \\
         \end{bmatrix}\Bigg).
\label{like3}
\end{align}
For inference using \cite{hensman2013gaussian} we need an expression for the conditional probability of the observations $\bm{y}$ given the latent calibration $\bm{f}$. So we need to perform a change of variable, to covert from the distribution of $p(\bm{\phi}(\bm{y},\bm{f}))$ to $p(\bm{y} | \bm{f})$. To do this we multiply the probability of a vector of calibrated observations, $\bm{\phi}$, by the determinant of the vector's Jacobian with respect to $\bm{y}$ (which is diagonal in this case), evaluated at the two observations.
\newcommand{\calfnz}[1]{\phi \big(z_#1,\f{#1} \big)}
\begin{equation}p\Bigg(
\begin{matrix}
y_i^{(1)} \\
y_i^{(2)}
\end{matrix} \Bigg|
\bm{F}
\Bigg)
 = 
 p\Bigg(\begin{matrix}
           \calfn{1} \\
           \calfn{2} \\
         \end{matrix}\Bigg)
\times
\frac{\partial \calfnz{1}}{\partial z_1} \Bigg|_{z_1 = y_i^{(1)}} \times \frac{\partial \calfnz{2}}{\partial z_2} \Bigg|_{z_2 = y_i^{(2)}}
\label{likelihood}
\end{equation}
For implementation, we use the log probabilities, and so the product becomes the sum of the log probability of $\bm{\phi}$ and the log of the two partial derivatives.
%

\textbf{Reference Instruments}
We define a binary vector $\bm{r}$, which describes which sensors are reference sensors. In those cases, we use the identity function instead of the calibration function, i.e. replacing $\calfn{a}$ with $y_i^{(a)}$.
\subsubsection{Variational inference for the calibration pair model}
We wish to compute the calibration parameters from the observations, $p(\bm{F}|\bm{Y})$. A simple application of Bayes rule, $p(\bm{F}|\bm{Y}) = {p(\bm{Y}|\bm{F}) p(\bm{F})}/{p(\bm{Y})}$, is not soluble or tractable in closed form,
%
as $N$ may be large and the likelihood function is non-Gaussian.
We quickly step through the variational approach, largely based on \citep{hensman2013gaussian}. 
We approximate the posterior, $p(\bm{F}|\bm{Y})$ with a variational distribution over $\bm{F}$: $q(\bm{F})$. We want to make this distribution similar to the true posterior. To this end we aim to minimise the KL divergence between the two:
\begin{align} 
\mathcal{D}_{KL}\Big[q(\bm{F}) || p(\bm{F}|\bm{Y}) \Big] &= - \int{q(\bm{F}) \log \frac{p(\bm{F}|\bm{Y})}{q(\bm{F})} d\bm{F}}\\
&= - \int{q(\bm{F}) \log \frac{p(\bm{F},\bm{Y})}{p(\bm{Y})q(\bm{F})} d\bm{F}}\\
&= - \int{q(\bm{F}) \log \frac{p(\bm{F},\bm{Y})}{q(\bm{F})} d\bm{F}} + \int{q(\bm{F}) \log p(\bm{Y}) d\bm{F}}\\
&= - \underbrace{\int{q(\bm{F}) \log \frac{p(\bm{F},\bm{Y})}{q(\bm{F})} d\bm{F}}}_{\text{The ELBO}} + \log p(\bm{Y})
\end{align}
The last term is constant wrt the variational distribution so we can minimise the KL divergence simply by maximimising the integral (the evidence lower bound,  ELBO). This is still not tractable without the approximation provided by inducing points. 
We introduce an additional vector $\bm{u}$ of values that we assume can describe $\bm{F}$, evaluated at inducing point locations, $\bm{Z}$. Each row of $\bm{Z}$ has a time, and a sensor id/parameter id, to index the associated GP in $\bm{F}$. We found that spacing the inducing points evenly over the sensor domains, rather than being optimised works well for the datasets we experimented with. 
We can compute the posterior over both $\bm{F}$ and $\bm{u}$ (we assume that $\bm{Y}$ is conditionally independent of $\bm{u}$ given $\bm{F}$),
$p(\bm{F},\bm{u}|\bm{Y}) \propto p(\bm{Y}|\bm{F}) p(\bm{F}|\bm{u}) p(\bm{u}).$ 
We don't compute this directly, but instead use our approximation,
$q(\bm{F},\bm{u}|\bm{Y}) = p(\bm{F}|\bm{u})q(\bm{u}).$ Here we have assumed that $\bm{u}$ is a sufficient statistic to determine $\bm{F}$ so that $p(\bm{F}|\bm{u},\bm{Y})=p(\bm{F}|\bm{u})$.
We substitute this approximation into the ELBO and augment with the inducing point values,
\begin{align}
\mathcal{L} &= \int\int{q(\bm{F},\bm{u}) \log \frac{p(\bm{F},\bm{u},\bm{Y})}{q(\bm{F},\bm{u})} d\bm{F} d\bm{u}}\\
&=\int\int{q(\bm{F},\bm{u}) \log \frac{p(\bm{Y}|\bm{F})\cancel{p(\bm{F}|\bm{u})}p(\bm{u})}{\cancel{p(\bm{F}|\bm{u})}q(\bm{u})} d\bm{F} d\bm{u}}\label{deriveelbo}\\
&=\int\int{q(\bm{F},\bm{u}) \log p(\bm{Y}|\bm{F}) d\bm{F} d\bm{u}} + \int\int{q(\bm{F},\bm{u}) \log \frac{p(\bm{u})}{q(\bm{u})} \; d\bm{u}\; d\bm{F}}
\end{align}
The second term's integral over $\bm{F}$ integrates out, leaving:
\begin{align}
\mathcal{L} &=\int\int{q(\bm{F},\bm{u}) \log p(\bm{Y}|\bm{F}) d\bm{F} d\bm{u}} + \int{q(\bm{u}) \log \frac{p(\bm{u})}{q(\bm{u})} d\bm{u}}\\
&=\mathbb{E}_{q(\bm{F},\bm{u})}\Big[\log p(\bm{Y}|\bm{F}) \Big] - \mathcal{D}_{KL}\Big[q(\bm{u})||p(\bm{u})\Big]
\end{align}
We now maximise $\mathcal{L}$ wrt the parameters of the variational distribution, described by the mean, $\bm{m}$ and triangular matrix $\bm{R}$ (covariance = $\bm{R} \bm{R}^\top$), using stochastic gradient descent. Specifically we approximate the expectation by sampling from $q(\bm{F})$ and then computing the mean of the log likelihoods for all the samples. We use TensorFlow's automatic differentiation to maximise $\mathcal{L}$ wrt the variational parameters, $\bm{m}$ and $\bm{R}$ using this stochastic approximation to the true gradient (stochastic gradient descent). This formulation also allows us to use minibatches, if $N$ becomes intractably large. See Algorithm \ref{vialg}, in the Supplementary Material for the overview of the computation.
Finally we can use the estimates of the variational parameters to predict $q(\bm{F})$ for a test point at time $t_*$ for all sensors.

\subsubsection{Optimising hyperparameters}
\label{optimise}
To avoid excessive model complexity we used just four hyperparameters: the lengthscale of static sensors; the lengthscale of mobile sensors; likelihood Gaussian noise; and the scale for all kernels.
One could treat these as random variables and integrate over them as with other variables. However, a common approach when using Gaussian processes is to find a point estimate either by maximising the marginal likelihood, or in our case, by optimising an error metric in held out data. We used Bayesian optimisation \citep[using the GPyOpt library, ][]{gpyopt2016} to find such a point estimate, which we used in a model tested on new data. Although we are not now reflecting the uncertainty in the hyperparameters, this point estimate means that the model when deployed in a data processing pipeline, is quicker and more robust. More importantly our lengthscale can reflect our prior belief that a sensor might degrade quickly \emph{even if it hasn't yet done so yet in the training data}. If such a failure doesn't yet exist in the dataset, the model is liable to select very long lengthscales. 
By manually selecting lengthscales we can incorporate our domain knowledge in sensor quality and performance.

\subsubsection{Sampling}

For the synthetic data example below, we found the algorithm typically optimised the calibration \emph{between} low-cost sensors first, then would very slowly move these distribution means to match the reference sensors. This behaviour is inevitable due to the highly correlated posterior and the imbalance in samples (with most being between non-reference pairs of instruments). Many elements of the approximating distribution mean must all move together in precisely the right direction, to increase the fit to the reference sensor data. As a side note, this might suggest parametrisations of the approximating distribution's mean vector should be investigated that might allow this manipulation using fewer variables. 
This problem is compounded by the relatively infrequent reference sensor observations: Many mini-batches contain few or no samples from the reference instrument co-locations. The gradient from the reference sensors only directly applies to those sensors that were co-located with it, while the co-locations of these sensors with the other non-reference collection of sensors will exert a gradient in the opposing direction.

We found for some experiments that importance sampling \citep{csiba2018importance} largely solved this issue and led to fast, reliable optimisation. We simply oversampled co-locations involving the reference sensors and corrected for the biased sampling by adjusting the noise variance appropriately.

We did not find this a problem in the experiment with real data from Kampala. This might be due to the shallow nature of the network, making it easier for the reference co-locations to influence the variational distribution across the whole network.
Part of our preprocessing of the real data involved averaging samples of co-located observations (blocks of 10 observations, typically over 15 minutes, were averaged), such averaging is standard practice in the field \citep[e.g.][]{stedman2006review}. This is necessary due to the considerable noise associated with individual observations (each averaging over only 80 seconds). This may have led to some balancing between the reference-sensors (which make measurements averaging over longer periods) and other sensors, which could explain why importance sampling was unnecessary for the real dataset.
\subsubsection{Other variables}
OPC PM2.5 sensors are sensitive to environmental factors. Specifically, a high relative humidity can lead to large biases \citep[e.g.][in particular figure 6]{crilley2018evaluation}. In this framework such variables could be included by simply adding additional columns to $\bm{Y}$. These can then be passed to the calibration function, $\phi$, and involved in the parametric expression that computes the `true' pollution.

\subsection{Calibration over categorical data}
To demonstrate the flexibility of the approach, we apply the method to another challenge from our research group: How to combine (low accuracy) citizen-science species-labelling of bee videos to predict the true species (with uncertainty quantification). Some of the videos were also labelled by an experienced and trained researcher, which we could consider ground-truth. Can we use the same calibration approach as above, treating each citizen-scientist as a low-cost sensor and the experienced researcher as a reference sensor? Previously (for sensor $j$) we were modelling the parameters, $\bm{f}_j(t_i)$ of a calibration function, $\phi(y,\bm{f}_j(t_i))$, between measured and true pollution - for example the gradient and offset. In the bee-labelling case, the measured observations are the labels guessed by the citizen scientists. The parameters we model are the probabilities in a conditional probability confusion matrix. These are generated using a softmax function applied to a matrix of latent variables which are given Gaussian process priors, allowing each citizen scientist's labelling to vary over time. Maybe people improve over time, with practice, or forget how to distinguish species over the winter (when fewer bees are foraging).

Previously the two observations $y_i^{(1)}$ and $y_i^{(2)}$ were taken at the same time $t_i$. In this categorical example we may need two times (one for each person) - as they may have performed the labelling in different orders and at different times, $t_i^{(1)}$ and $t_i^{(2)}$.

To switch to a categorical dataset, the only part of the algorithm that needs to change is our likelihood function defined previously in \eqref{likelihood}.
\newcommand{\species}{\psi}
\newcommand{\numspecies}{A}
With number of species = $\numspecies$, we can construct a conditional confusion matrix, $\bm{P}$, using $|\bm{f}_j| = \numspecies^2$ parameters. Each element, $P_{y,\species} = p(y|\species,\bm{f}_j)$, of $\bm{P}$ describes the probability of the observed result \emph{given} the real bee was of (latent) species, $\species$. We build this matrix from the vector of parameters, $\bm{f}_j$, by simply reshaping the vector into a square $\numspecies \times \numspecies$ matrix,  $\bm{C}$, and use the softmax function to normalise each column, so that $\sum_y P_{y,\species} = 1$: \begin{equation}P_{y,\species} = e^{{C}_{y,\species}(t)} / \sum_z e^{{C}_{z,\species}(t)}.\end{equation}

The likelihood function takes two observations of the same bee, $y_{i}^{(1)}$ and $y_{i}^{(2)}$ and two parameter vectors $\ft{1}$ and $\ft{2}$, where $s_{i}^{(1)}$ and $s_{i}^{(2)}$ are the identities of the sensors (citizen scientists).
We compute $\bm{P}$ for the two sensors/citizen scientists involved and take the relevant column from each, to get $p(y_{i}^{(1)}|\species, \ft{1})$ 
and $p(y_{i}^{(2)}|\species, \ft{2})$ 
We multiply these together, assuming that there is conditional independence between the citizen scientists, to compute the conditional joint probability, \begin{equation}p(y_{i}^{(1)},y_{i}^{(2)}|\species, \ft{1}, \ft{2}) = p(y_{i}^{(1)}|\species, \ft{1}) \; p(y_{i}^{(2)}|\species, \ft{2}).\end{equation}
Using either the training data, or wider ecological literature, we define a categorical prior distribution over the true species $p(\species)$. We multiply the conditional joint probability vector with this prior and sum over the resulting matrix, marginalising out the latent species:
\begin{equation}
p\Big(y_{i}^{(1)},y_{i}^{(2)}|\ft{1},\ft{2}\Big) = \sum_\species p\left((y_{i}^{(1)}|\species, \ft{1}\right)  \; p\left(y_{i}^{(2)}|\species, \ft{2}\right) \; p\Big(\species\Big).
\end{equation}

As before, we sample values of $\bm{f}$ from the variational approximation, $q(\bm{f})$, and compute a Monte Carlo approximation to the likelihood. An autodiff framework, such as tensorflow, again allows us to optimise the variational parameters. In the implementation the only component that changed from the air pollution example was the method for computing the likelihood.

\subsection{Multi-hop instant calibration}
\label{hasenfratz}
For comparison with our calibration pair method, for the air pollution examples we developed and implemented an algorithm loosely based on the \emph{Multi-hop Instant Calibration} method from \cite{hasenfratz2012fly} and the \emph{rendezvous connection graph} from \cite{saukh2015reducing}, the former paper is a little unclear about combining weights, while the latter one handles sensor drift by only using observations within each time window. In \cite{saukh2015reducing} a shortest path algorithm is applied (personal communication with authors) to convert the network to an acyclic graph.

\begin{figure}[t!]
  \centering    
    \includegraphics[width=0.3\columnwidth]{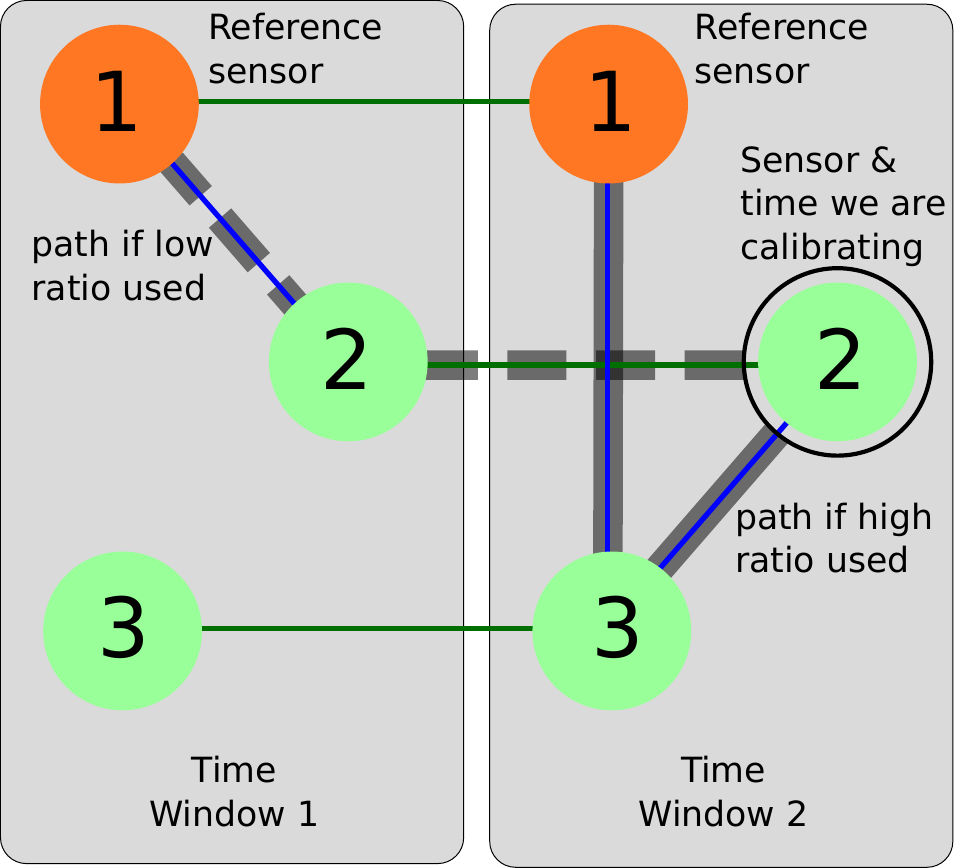}
  \caption{Multi-hop: The ratio of edge weights for neighbouring times (green) over weights for colocated sensors (blue) affects the choice of shortest path to a reference instrument. A low ratio will lead to shorter chains of colocation but relying on colocations from different time windows (dashed grey path), a high ratio will lead to a preference for colocations from nearby times, but with longer chains of colocations pairs (solid grey path).}
    \vspace{-3mm}    
  \label{graphidea}
\end{figure}

We augment the original graph in \cite{saukh2015reducing} by adding edges between the same sensor, in neighbouring time windows. The intuition being that the scaling information provided by a long series of hops within the window might be less accurate than a direct reference calibration `carried over' from a different time window. The edges connecting time windows can have a different weight than those connecting sensors in the same time period. The choice of weight ratio (i.e. time-edge-weight/colocation-edge-weight) will decide whether the model leans towards using longer chains of colocations in the same time window (high ratio) vs relying on colocations that occurred a long time ago (low ratio). Figure \ref{graphidea} illustrates.
We consider the calibration to be a simple scaling factor in the experiments in this paper and so we restrict the calibration function in both the multi-hop and in our more complex variational inference calibration tool to also scaling-only when performing comparisons.

The implementation of the graph building and prediction are detailed in Algorithms \ref{mhga} and \ref{mhgpa} respectively, in the Supplementary Material.

\section{Results}


\subsection{Synthetic example}

\subsubsection{The synthetic data}

Before considering real data we explored the method's response to a representative, but synthetic, dataset. We simulated ten static and four mobile sensors over a 180 day period. Four of the static sensors are assumed to provide noise-free, unbiased, reference measurements, the rest have a time-varying bias and added Gaussian white-noise with variance of $100 \;(\mu \text{g/m}^3)^2$. In the Kampala data discussed later, each mobile sensor is typically localised to one part of the city, with only occasional visits further afield. We simulate this by making the probability of a visit by a mobile sensor to a static sensor, proportional to the inverse distance between a `home' and the static sensor. Simple sinusoidal functions were used to generate both the true pollution and the scaling of the mobile and static sensors. The period of the sinusoids was different for each sensor, but the static sensors had a longer period on average (median = $3070 \;\text{hours}$ vs $1007 \;\text{hours}$). This variation was to (a) emphasise the heterogeneous nature of the sensors and (b) illustrate how the mobile sensors are probably less stable. Figure \ref{results_synth} shows the scalings actually used.


\subsubsection{Variational pair model}
We applied the method described in Section \ref{calpair} to the synthetic data. As mentioned in Section \ref{optimise}, we used Bayesian optimisation (30 iterations, optimising the NMSE) to select the model hyperparameters for the variational calibration pair method (kernel variance, likelihood variance and two lengthscales). We optimised using the synthetic data with added noise scale of $10 \mu g m^{-3}$ for all the results. The synthetic data was resampled each iteration of the Bayesian optimisation (and for testing): the sensor locations, colocation events and noise were all resampled; but the frequencies of the synthetic calibration sinusoids remained the same.

Even though each sensor's scaling function has a different period, we use just two lengthscales to model the static and mobile sensors. On the first dataset generated, the Bayesian optimisation chose $2201 \;\text{hours}$ and $581\;\text{hours}$ for the lengthscales of the static and mobile sensors, respectively. This likely reflects both the underlying generating function, but also the strength of evidence for shorter lengthscales. The longer lengthscales assigned to the static sensors is longer than the period of the median sinusoid, but it potentially allows those static sensors with a longer lengthscale to still be modelled well, by allowing more data to be used for any given prediction. It also permits calibration information to be passed over longer periods of time.
The likelihood variance and kernel scale (variance) were $6.02 \;(\mu \text{g/m}^3)^2$ and $7.64 \;(\mu \text{g/m}^3)^2$ respectively. These reflect the variance of the scaling factor, so it is hard to compare directly to the noise added to the data. The choice of likelihood variance seems to overestimate the true value a little, and it may be that the longer lengthscales chosen for the static sensors have their inaccuracy `explained away' by this added noise term.


\begin{figure}[t!]
\includegraphics[width=\textwidth,trim=100 0 100 0]{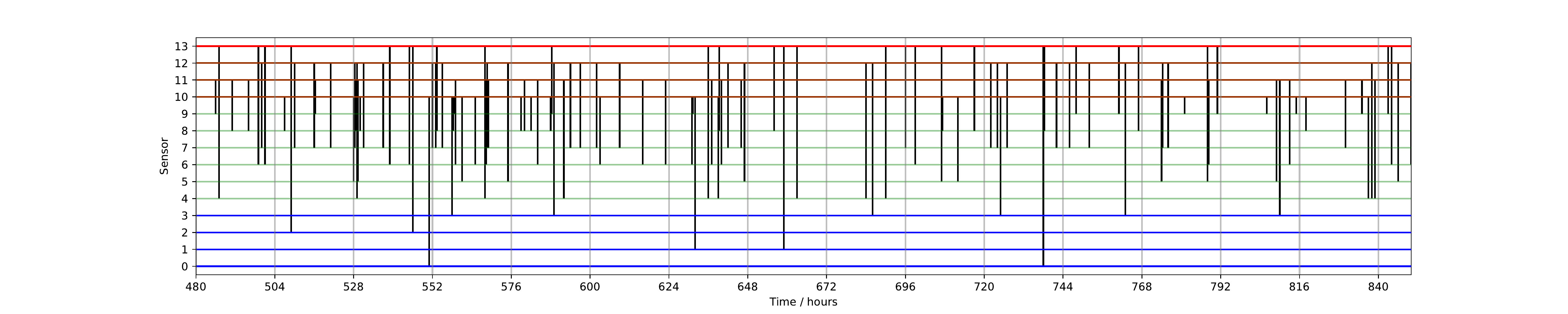}
  \caption{Synthetic sensor example: Colocation visits over a representative 350 hours of the total 4320 hours: Visits to the reference (blue) and low-cost (green) static sensors by the mobile ones (red), indicate by vertical black lines.}
    \vspace{-3mm}    
  \label{synthetic_sensorplacement}
\end{figure} 
%
\begin{table}
\centering
\begin{tabular}{c |c c c| c} 
Noise Scale / $\mu g\; m{}^{-3}$ & \multicolumn{3}{c}{NMSE} & NLPD \\
 & No Method & Multi-hop & Cal. Pair Model & Cal. Pair Model \\
2 & 0.77 (0.02) & 0.27 (0.01) & 0.18 (0.01) & 42.6 (2.8) \\
5 & 0.78 (0.02) & 0.44 (0.03) & 0.21 (0.01) & 49.5 (2.3) \\
10 & 0.97 (0.02) & 0.87 (0.07) & 0.35 (0.01) & 58.2 (2.3) \\
20 & 1.46 (0.02) & 3.18 (0.69) & 0.84 (0.02) & 87.7 (1.5) \\
\end{tabular}
\caption{NMSE using raw data, the multi-hop calibration and the variational calibration pair model, for synthetic data with four levels of added noise. Bracketed values are the standard error of the mean based on predictions generated from 10 synthetic datasets (for each, the multi-hop was optimised on the test data, while the calibration pair model was optimised once on training data with noise scale = 10).}
\label{synth_table}
\end{table}

\subsubsection{Multi-hop calibration and comparison}
We also applied the multi-hop calibration method to the synthetic data. We first performed a grid search to optimise the window size and the ratio of edge weights between those edges representing time and those representing a colocation (again, performed on separately generated training data). 
 We unfortunately found this model performed very poorly on this data. The best configurations usually had very long window sizes, leading to little or no ability to model the variation over time. We found that this configuration marginally improved on simply using the raw measurements. See Table \ref{synth_table}.

\begin{figure}[t!]
  \centering    
   \begin{subfigure}[b]{0.44\textwidth}    
    \includegraphics[width=\textwidth,trim=60 40 30 0]{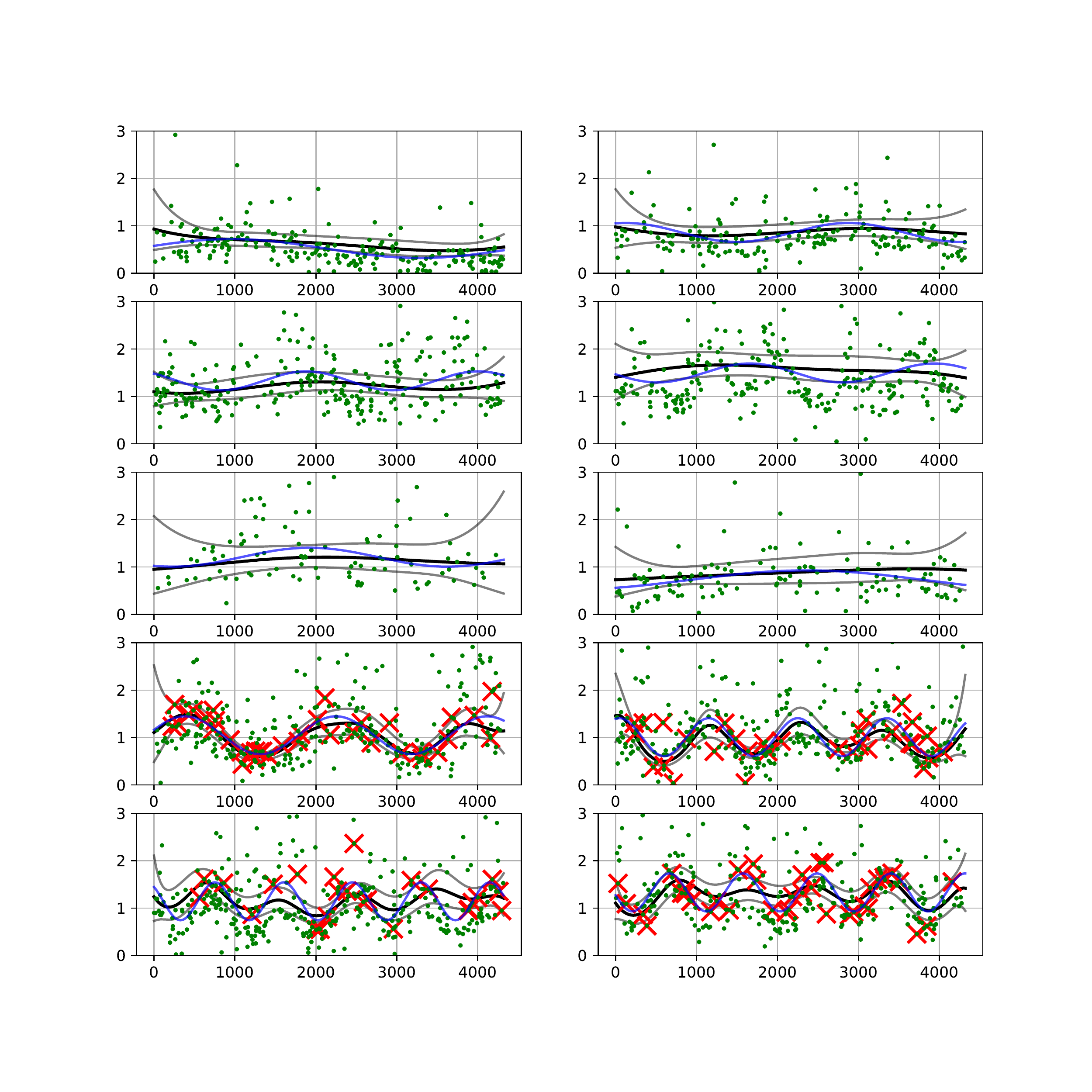}
     \caption{Variational Calibration Pair Model}
   \end{subfigure}
   \begin{subfigure}[b]{0.44\textwidth}    
    \includegraphics[width=\textwidth,trim=30 40 60 0]{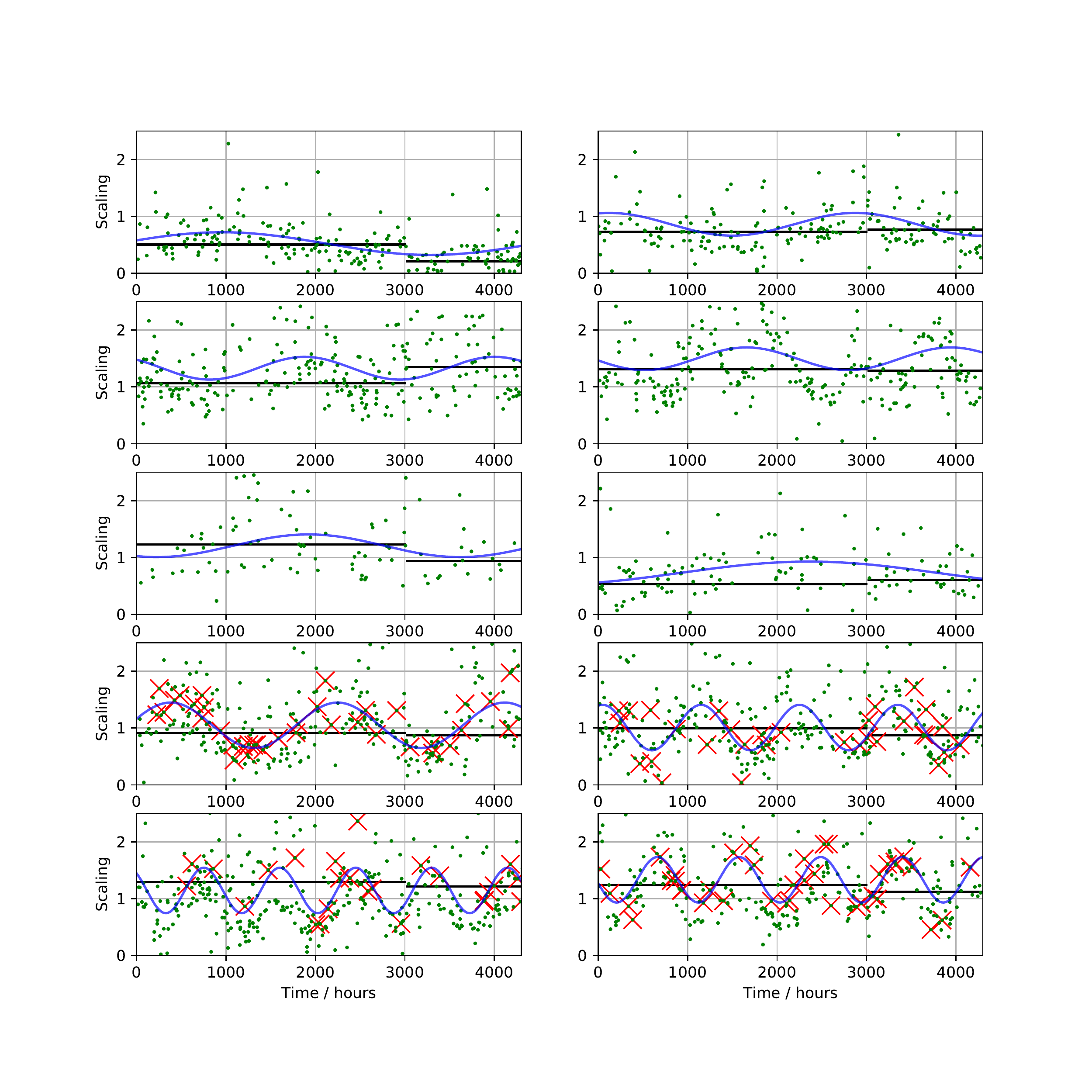}    
     \caption{Multi-hop}
   \end{subfigure}

  \caption{Predictions for the scaling in the synthetic sensor example for both the Variational Pair Model and the Multi-hop method. Noise scale = 10 $\mu g/m^3$. Upper six plots in each, static sensors. Lower four plots, mobile sensors. Black line, posterior mean. Grey lines, 95\% of the posterior density. Blue line, true scaling of each sensor. Green dots, ratio of colocation observation pairs involving this sensor. Red crosses, ratio of colocations observations between this sensor and a reference sensor.}
    \vspace{-3mm}    
  \label{results_synth}
\end{figure}

The estimates from the variational method provided considerable improvements over the multi-hop approach. 
And we can also compute the negative log predictive  as we predict a distribution. 
Figure \ref{results_synth} shows these predictions for the 6 non-reference static sensors and the four mobile, for both the multi-hop method and the calibration pair model. One can see some errors in the calibration pair model. In particular, the high frequency drift of the low-cost mobile sensor has not been properly detected by the model in all the low-cost sensors, leading to this drift appearing in the predictions for the other sensors.

The synthetic data was engineered specifically to provide low-quality, noisy, infrequent and challenging data, in an attempt to reflect the problems that are typical in real data. Figure \ref{synthetic_sensorplacement} illustrates how few visits were made to the reference instruments, over a representative sample of the synthetic data. To explore this further, we added different amounts of noise to the data, and reran the analysis, although we kept the variational calibration method's hyperparameters as those trained with a noise scale of 10 $\mu g\; m{}^{-3}$, to see how well it adapts to misspecified noise.
Table \ref{synth_table} reports the NMSE for synthetic data with four added noise-scales for the two methods. The multi-hop method is allowed to optimise its parameters for each noise scale and, as might be expected, its window size reduces as the noise is reduced (for noise scales of $2, 5, 10$ and $20$ $\mu g m^{-3}$ the selected window sizes were $292, 1184, 3010$ and $4800$ hours, respectively). 
We can also see that the accuracy of the multi-hop method approaches the calibration pair model's, as the noise is reduced. This is probably because the calibration pair model makes use of multiple pathways through multiple colocations and over time to support a prediction, whilst the multi-hop method is constrained to use a single path through the network for a given prediction, making it more vulnerable to the effect of noise. The use of multiple pathways in the pair model gives it considerable robustness against noisy data.
\subsection{Kampala air pollution data}
\label{kampaladata}
\begin{figure}[t!]
  \centering    
    \includegraphics[width=0.8\columnwidth,trim=55 55 55 55,clip]{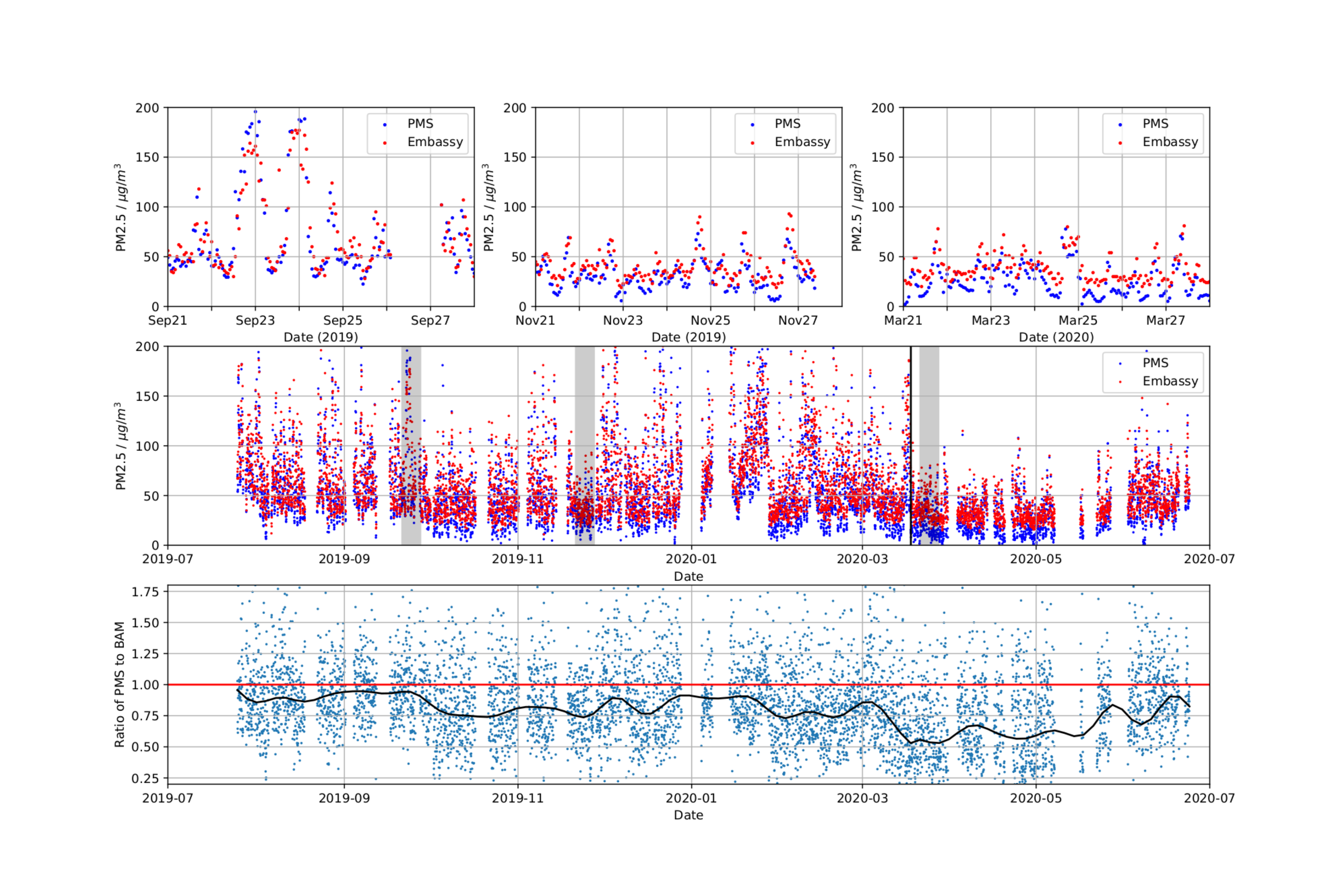}
  \caption{Upper plots, enlargements of three weeks (highlighted in grey in the middle plot), showing correlation between the low cost sensor and the reference instrument. Middle plot, the full period of measurements. Discontinuity on 18th March, 2020, due to the covid-19 lock down (vertical black line). Lower plot, the ratio of the two sensors. A GP with RBF kernel has been fitted to the log ratio and the exponent of its posterior mean plotted (pre- and post- lock down modelled with independent GPs). Times are in UTC.}
    \vspace{-3mm}    
  \label{embassy_ratio}
\end{figure}
\begin{figure}[t!]
  \centering    
    \begin{subfigure}[b]{0.44\textwidth}    
    \includegraphics[width=\textwidth]{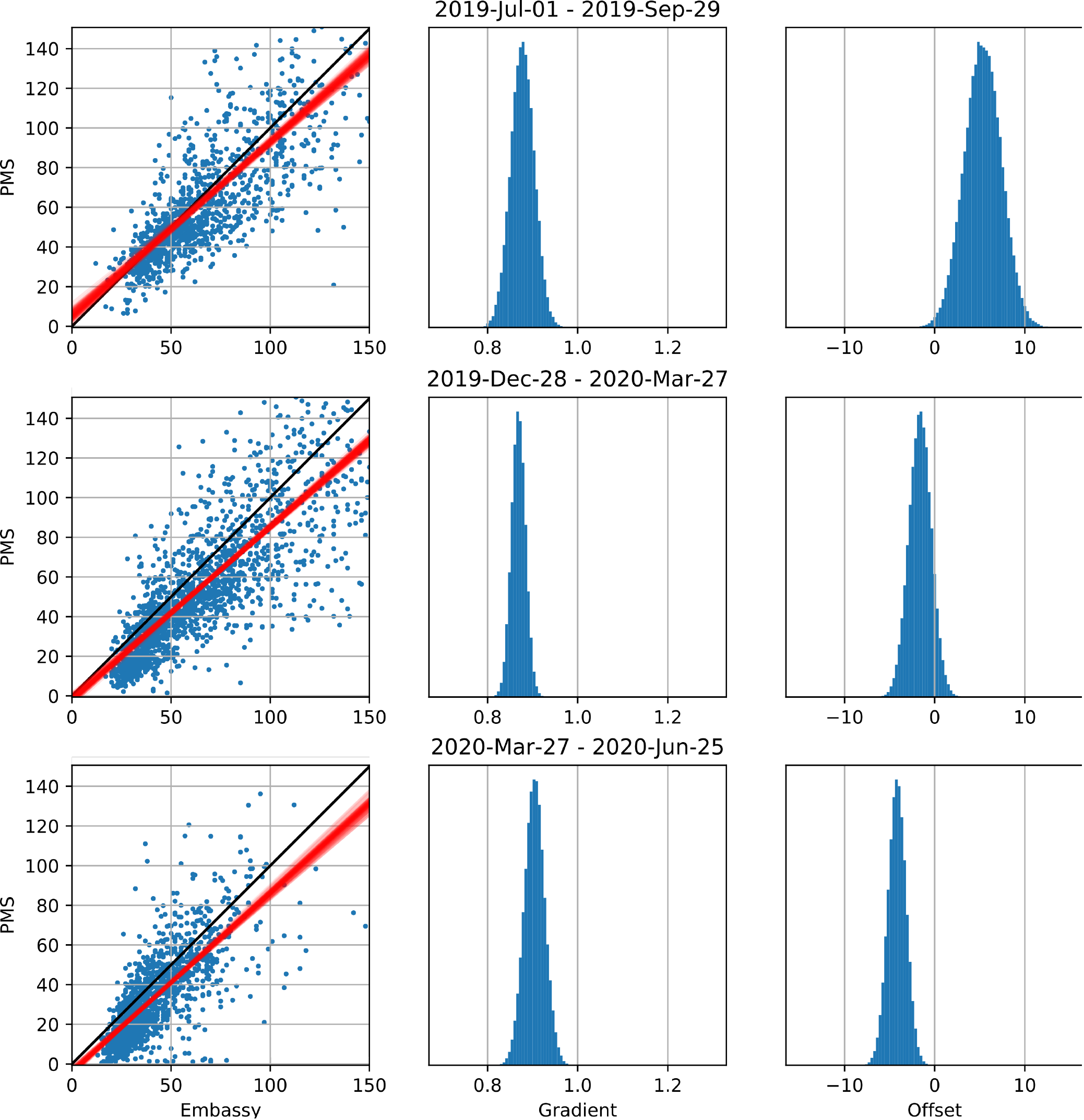}
    \caption{Static OPC measurements vs the embassy data}    
    \end{subfigure}
    \begin{subfigure}[b]{0.44\textwidth} 
    \hspace{0.5cm}   
    \includegraphics[width=\textwidth,trim = 6cm 1.5cm 7cm 0,clip]{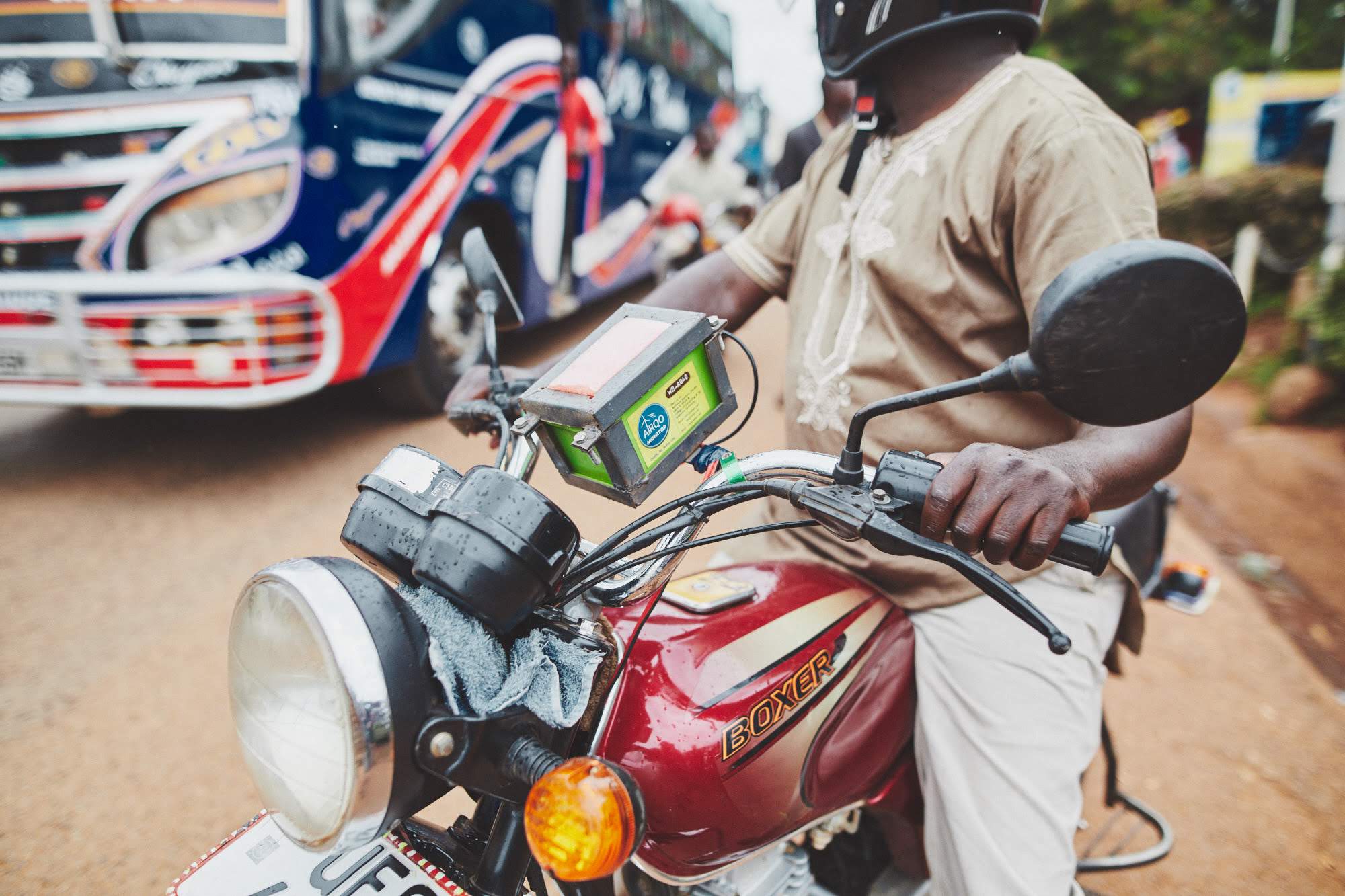}
    \vspace{0.4cm}	
    \caption{A mobile OPC}
    \end{subfigure}    
  \caption{(left) OPC measurements vs the embassy data. Each row corresponds to 90 days of measurements. The left scatter plot shows the raw connection between the two sensors. A black line indicates equality while red lines are sampled from the linear fit's posterior probability distribution. Middle and right plots are the gradient and offset marginal posterior distributions respectively. The gradient distribution remains roughly constant, while the offset appears to  `drift' downwards over time. (right) One of the mobile low-cost sensors mounted on the front of a motorbike taxis (`boda boda'), photographed in Kampala by AirQo (\url{www.airqo.net}).}
    \vspace{-3mm}    
  \label{embassydemo}
\end{figure}

 
Recently a network of low-cost particulate air pollution sensors has been deployed across Kampala. These consist of pairs of PMS5003 (Plantower) optical particle counters (OPCs) mounted at fixed, static locations and on motorbike taxis, known as boda bodas (Figure \ref{embassydemo}b). Before discussing the network we will briefly inspect the measurements of a static OPC mounted 10m from the US Embassy's air pollution monitor (a regularly calibrated BAM sensor, which provides hourly ground truth measurements of PM2.5 pollution). This brief comparison will give some insight into the calibration problem: What the calibration function might consist of, and how it might vary over time.

Figure \ref{embassy_ratio} illustrates the strong correlation between the two sensors, but also highlights some differences. The earlier plot from September seems to show very similar measurements, with a ratio near one, on average. But later measurements seem to have a greater discrepancy with the low cost OPC sensor significantly underestimating the true pollution. The ratio itself is not enough to explain this change. Figure \ref{embassydemo}a illustrates how an offset can explain some of the difference. In all three the gradient of the linear fit appears to remain between 0.85 and 0.95, however the offset maybe drifts from positive to negative, explaining the increasing disparity in the previous graphs.

\subsection{Calibration over the network}

\begin{figure}[t!]
  \centering    
   \begin{subfigure}[b]{0.27\textwidth}      
    \includegraphics[width=\textwidth,trim=0 0 0 0]{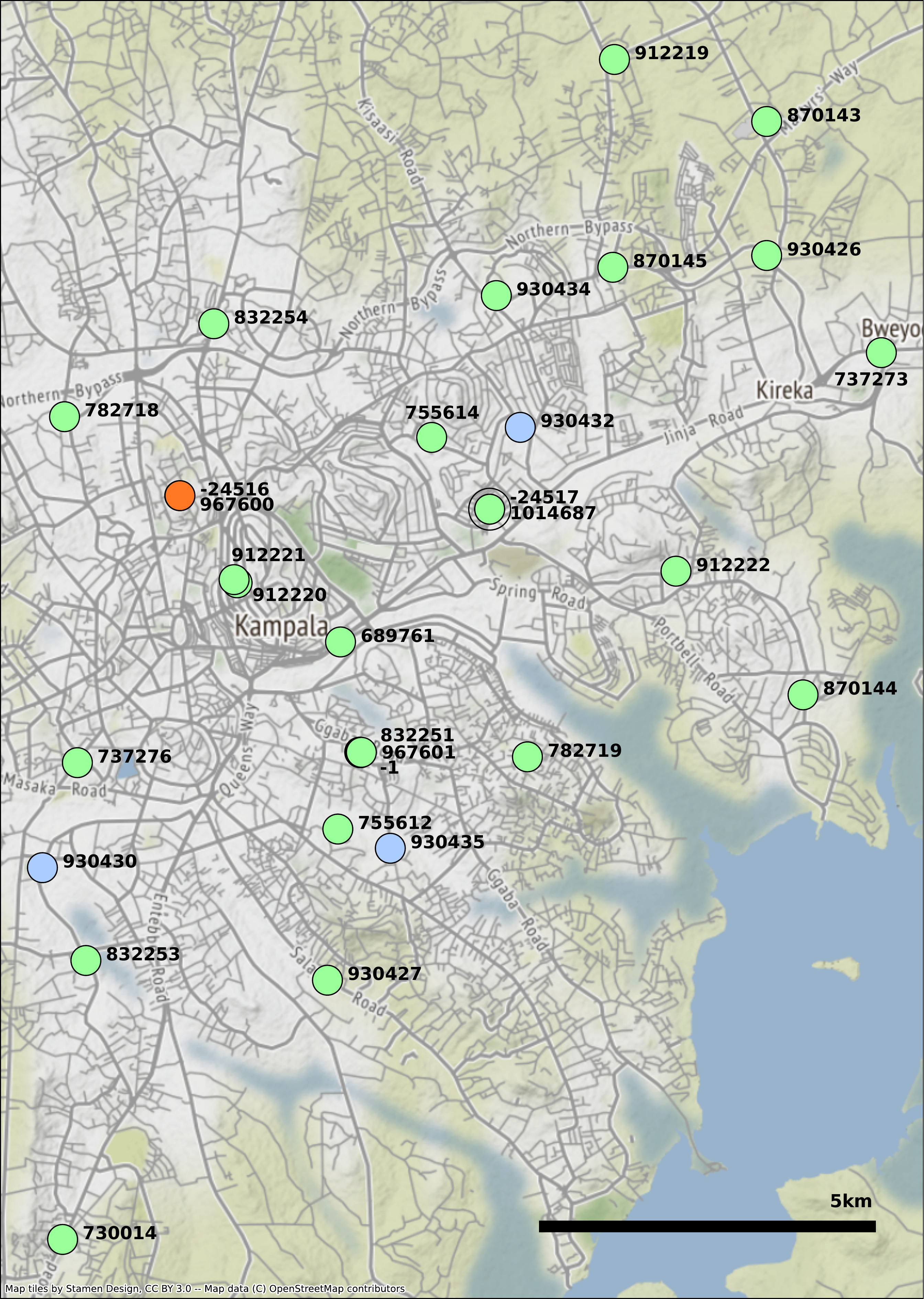}
     \caption{Kampala with sensor locations}
   \end{subfigure}
   \begin{subfigure}[b]{0.47\textwidth}   
    \hspace{0.5cm}  
    \includegraphics[width=\textwidth,trim=0 0 0 0]{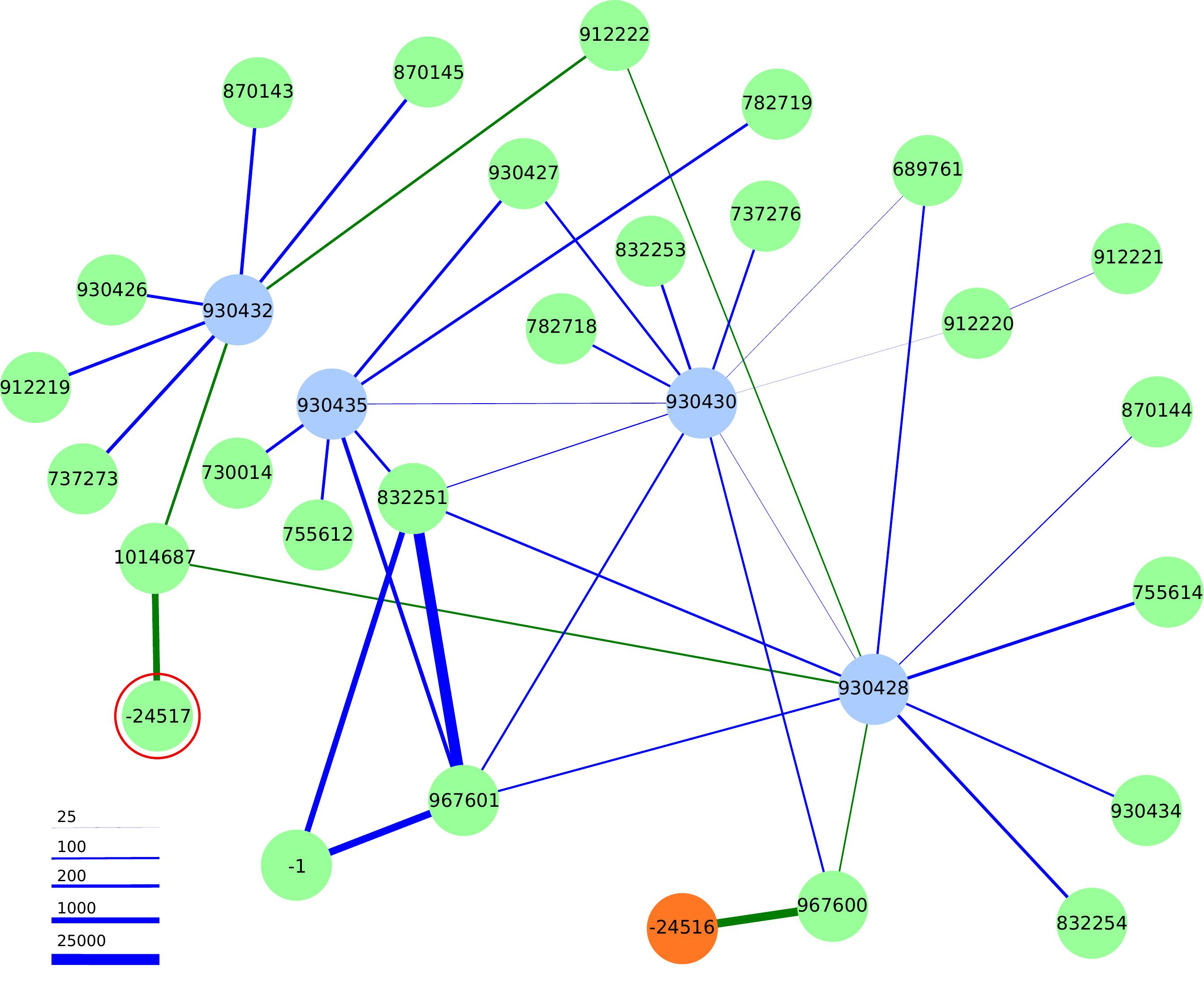}
     \caption{Colocation frequency graph}
   \end{subfigure}
    
  \caption{The AirQo air pollution monitoring network (sensors that have been colocated). The four blue nodes are the mobile sensors. The orange node is the reference sensor at Makerere University, the circled node is the reference sensor whose calibration we are testing. (a) Locations of sensors, the latest locations of mobile sensors are plotted, with one not in the city limits. (b) The network of colocations between sensors (from July 2020 to February 2021). The line thicknesses indicates the number of colocations in that time. The green lines indicate two possible paths from the reference to the test sensor.}
    \vspace{-3mm}    
  \label{network1kampala}
\end{figure} 
51 sensors were in the network in the second-half of 2020, consisting of 2 reference sensors (BAM 1022 Particulate Monitor, Met One), 4 mobile OPCs and 45 static OPCs (PMS5003, plantower). Between the 15th July, 2020 and 3rd February, 2021 (203 days) there were 6,118,977 measurements of PM2.5 recorded. We use this dataset for the analysis.
We built a dataset of 433,935 `encounters', consisting of every pair of measurements that occurred within 40m and 30 minutes of each other. We averaged ten minute periods 
and removed observations outside the $10-300 \mu g/m^3$ range, as these we found to sometimes be associated with faults. This left 40,432 records (approximately 200/day). The majority of these are associated with permanently co-located pairs of static sensors. 
In Figure \ref{network1kampala}, (a) shows the physical locations of the sensors and (b) the number of colocations between them. The two reference sensors `-24516' (on Makrerere University Campus) and `-24517' (in Nakawa), are approximately 5km apart.

To test the models we only defined -24516 as being a reference sensor and let the models estimate the calibration required of -24517. We gave the models all the colocation data (none was held out) as we are interested in the prediction of the calibration scaling correction for the test sensor. There are co-location observations at the test sensor over a period of 69 days from the 203 days total.

We would expect the test sensor to have a calibration scaling of exactly 1.0 as it's a high quality, carefully maintained and calibrated BAM sensor. Our Gaussian process model has a prior mean of the log-ratio of zero, which is exactly a gain of 1.0, while the multi-hop model doesn't have a prior. To avoid our prior giving an unfair advantage and to simulate the effect of a drifting calibration, as is seen in the low-cost OPCs (Section \ref{kampaladata}), we artificially scaled the measurements of the reference sensor to slowly increase over time (starting at 1.0$\times$ on the first day of observations at the test sensor, and increasing to 1.83$\times$ by the 69th day). In summary: The algorithms being tested must reconstruct this drift using the network of colocations from the known reference sensor.
%
For our calibration pair model we select hyperparameter values that we believe are appropriate for this dataset (lengthscale for static and mobile sensors = 100 days, RBF kernel variance = 9 $(\log(\mu \text{g/m}^3))^2$, Bias kernel variance = 3 $(\log(\mu \text{g/m}^3))^2$, likelihood (ratio) noise variance = 0.2 $(\log(\mu \text{g/m}^3))^2$). This approach is necessary, as discussed previously, to allow us to include our prior knowledge around lengthscales associated with sensor drift and failure.
We again compare to the multihop method described in Section \ref{hasenfratz}. For comparison with our method, we optimised the parameters of the competing multi-hop method (the window size and weight ratio between time and colocation) on the dataset we also test with. Effectively giving an upper bound on its capability.

\begin{figure}[t!]
  \centering    
    \includegraphics[width=0.85\textwidth,trim=55 5 55 5]{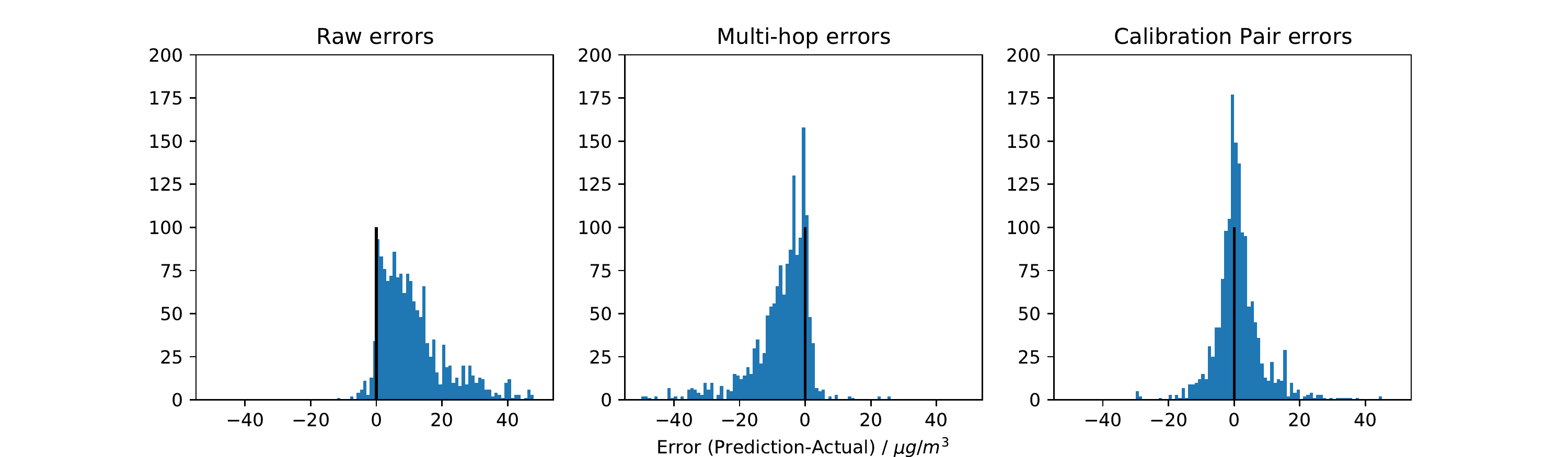}
  \caption{The distribution of errors in pollution predictions at the test sensor. As we give the three models the measured pollution, in principle the error could be zero if the model correctly estimates the calibration. The left graph is the distribution of predictions if we just use the measurements as the predictions. The added, synthetic, drift causes this to over-estimate. The central graph is using the multi-hop approach, this appears to underestimate. The right graph is for the calibration pair model. Runs of the variational inference optimisation end with slightly varying results here. This is a typical example (with a MAE about the average).}
    \vspace{-3mm}    
  \label{correcting_kampala}
\end{figure}

\textbf{Results} 
After running the grid search optimisation, the optimum multi-hop configuration had a window size of just 24 hours and an edge weight ratio of 0.21 (it roughly equally weights a colocation with a reference sensor from 5 days earlier and via an intermediate sensor during the same 24 hours). We computed error metrics between the test values of the reference sensor and the calibration predictions.
To review the aim: The two calibration methods used the other reference instrument and the network of colocations to estimate the calibration of the test instrument (with the added synthetic drift). Table \ref{kampala_results} outlines the results for these approaches, while Figure \ref{correcting_kampala} shows the distribution of errors from the different approaches. The `raw measurement' errors are caused by our artificial scaling (without this the raw measurements would have no error).
\begin{table}
\centering
\begin{tabular}{c |c c c | c} 
 Metric & No calibration & Multi-hop calibration & Calibration Pair Model & \\
 \hline
 MAE & 12.71 & 8.67 & 5.545 (0.47) & $\mu g\; m^{-3}$\\ 
 NMSE & 0.321 & 0.180 & 0.078 (0.011) & \\
\end{tabular}
\caption{Kampala experimental results, with synthetic drift added. Using no calibration, the multi-hop calibration and the pair model (bracketed values are standard error from 20 runs. Not required for multi-hop as it's deterministic).}
\label{kampala_results}
\end{table}
\begin{figure}[t!]
  \centering    
   \begin{subfigure}[b]{0.48\textwidth}        
    \includegraphics[width=\textwidth,trim=50 0 50 0]{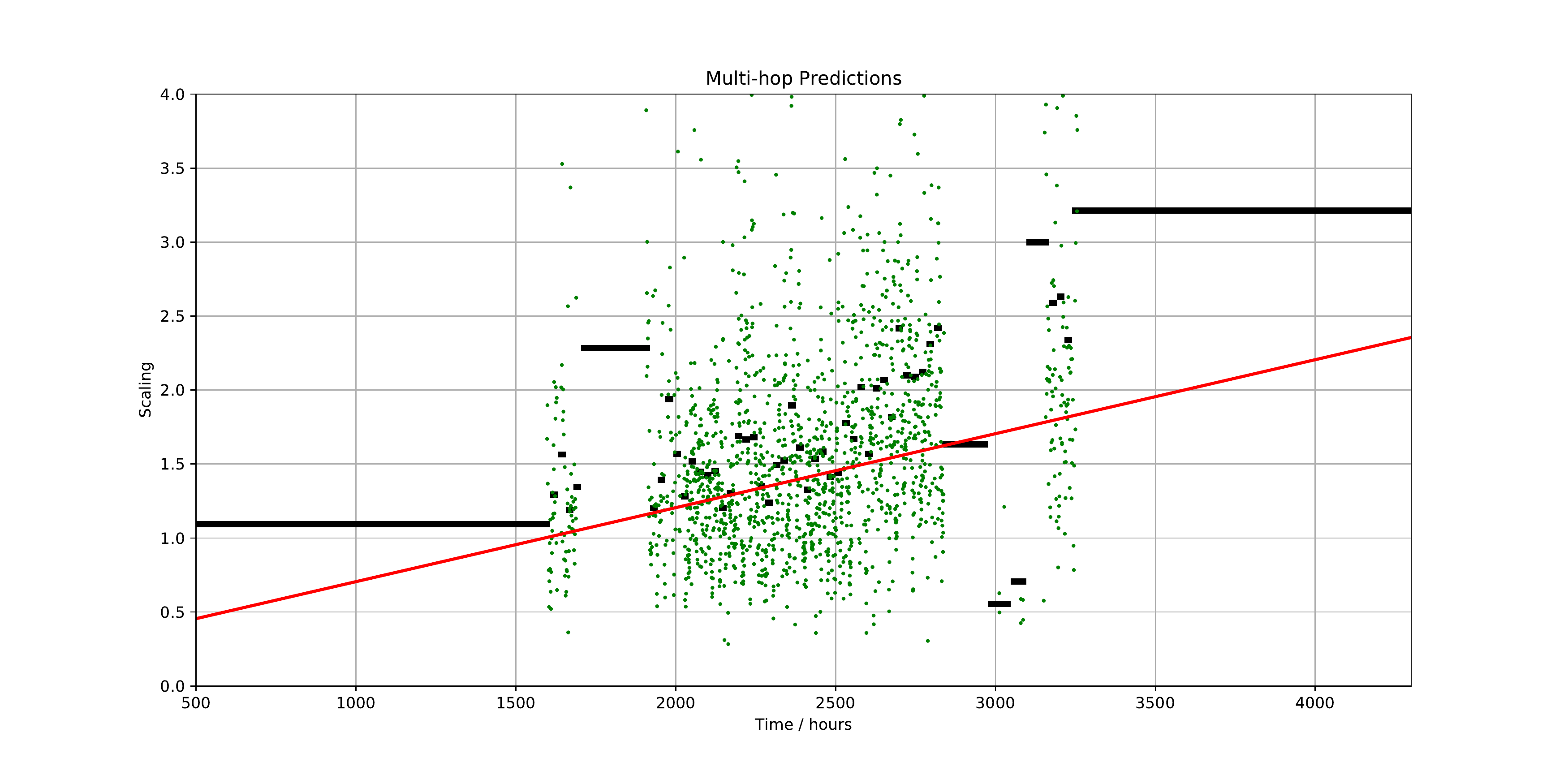}
     \caption{Multi-hop scale predictions}
   \end{subfigure}
   \begin{subfigure}[b]{0.48\textwidth}            
    \includegraphics[width=\textwidth,trim=50 0 50 0]{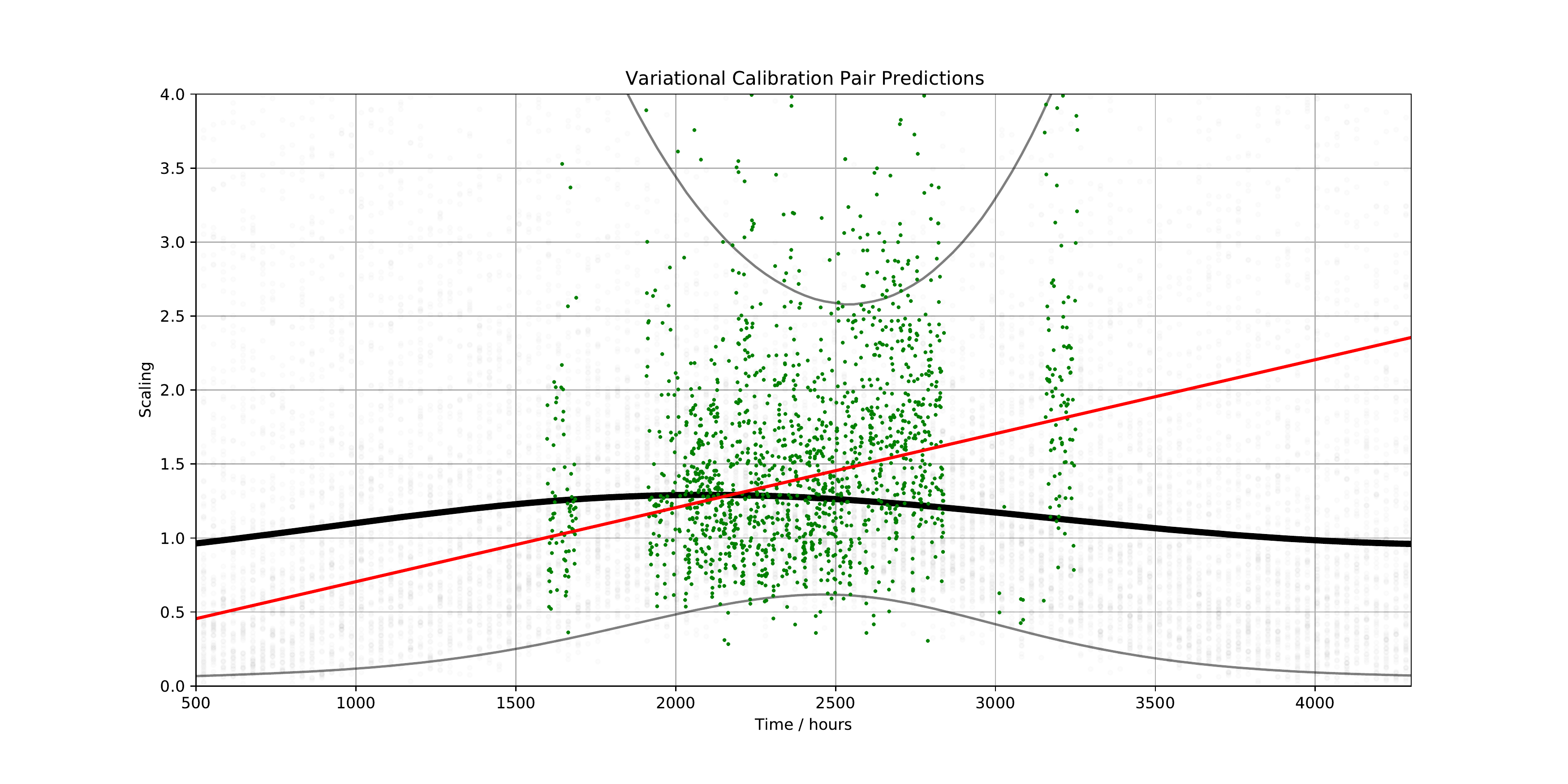}    
     \caption{Variational pair model predictions}
   \end{subfigure}
  \caption{The predictions of the scaling at the test sensor. Red line, our synthetically generated scaling drift we are trying to predict. Black line, the prediction. Green dots, ratios between the test sensor and other sensor(s) that visit. The variational prediction posterior mean varied quite a lot per run of the algorithm, probably due to high uncertainty, leaving the choice of mean difficult to determine.}
    \vspace{-3mm}    
  \label{timeseries_kampala}
\end{figure} 
We find the multi-hop method does help improve the accuracy of the predictions to an extent, but the variational method performs considerably better. Besides accuracy, the variational pair-model method provides uncertainty estimates for its predictions, and the associated 95\% confidence intervals include the true scaling function. Variational inference is associated with underestimates of the variance of the posterior \citep{blei2017variational} so this result is reassuring. 
The co-location data is very noisy. To illustrate this, the green markers in Figure \ref{timeseries_kampala} are the ratios between the test sensor and other sensor(s) that visit. Note that we are not assuming that these are just noisy samples from the true latent scaling - as these OPC sensors will themselves have some scaling associated with their estimates. But plotting them gives the reader a sense as to the scale of the noise in the dataset.

\textbf{Real Calibration} As a final brief experiment we tried calibrating the low cost OPC sensor (1014687) that is colocated with the test reference sensor, then using the calibration combined with the OPC observations to predict the values measured by the test sensor. The sensor was new and we only have 69 days of data associated. We found the actual scaling was the OPC was underestimating the reference sensor by only about 6\% which equates to an error of about 3 $\mu \text{g/m}^3$. The raw data had a MAE of 15.08 $\mu \text{g/m}^3$ while the multi-hop and variational pair model approaches had a MAE of 16.07 and 14.87 respectively. Neither approach really had much effect on the MAE. The earlier demonstration with the more significant artificial scaling was however modelled successfully, giving us confidence in the approach.
%
\subsection{Categorical data}
\subsubsection{Synthetic example}
We demonstrate the approach initially with a synthetic dataset of 300 bees, approximately evenly split between three species (36\%, 30\%, 34\%). We simulated three non-expert (NE) citizen scientists, and an expert (providing effectively the ground-truth reference). The NE capabilities vary over time, with NE A perfectly labelling the first half, then degrading to chance linearly over the remaining half. NE B starting at chance, improving linearly to perfect over the first half and providing perfect predictions for the last half. NE C distinguishes between classes \{1,2\} and 3 initially but, linearly, their classification can distinguish between \{1,3\} and 2 at half way, and then almost chance by the end. 

For training, 173 rows had access to the ground truth, leaving the algorithm to predict the remaining 127. The three NEs (A,B and C) made 178, 190, and 200 observations respectively.

To demonstrate we ran the algorithm using either an EQ kernel (scale=25, lengthscale=25\% of the length of the list of images) or a Bias kernel (scale=25). The scale and lengthscales were set \textit{a priori}. The result is a Gaussian process for each of the 27 latent function priors (nine in each confusion matrix for each of the three citizen scientists).
For comparison we used a collaborative filtering matrix factorisation approach, in which a binary matrix was first computed for each species (ones are where the species exists in the original dataset). We then used probabilistic matrix factorisation with bias offsets \citep{Hug2020} to predict the missing values in the binary matrices for the reference column, then assign the species to the matrix with the greatest value. We also computed more straightforward baselines: `most guessed' simply does voting, while `trust weighted' assigns a weight to each citizen scientist. The weights are optimised on the training data. We manually optimised the NLPD for these methods to give them the best chance of beating the calibration pair approach (e.g. we found adding 0.2 to the trust weighted sum of guesses for each class, minimised the NLPD for the synthetic data). We also tried weighting the votes by the prior (number of times each bee appears). We finally just report the most common.

Table \ref{cat_synth_table} shows both calibration pair methods achieve higher accuracy and better uncertainty quantification than the others, but the EQ kernel version substantially improves on this uncertainty estimation.

\begin{table}
\centering
\begin{tabular}{ c r r }
\hline
Method & Accuracy (\%) & NLPD \\
\hline
Calibration (EQ kernel) & 80\% & 53.5 \\
Calibration (Bias kernel) & 77\% & 78.1 \\
Collaborative Filtering & 53\% & 128.1 \\
Most Guessed & 74\% & 81.9 \\
Most Guessed (`trust' weighted) & 76\% & 81.7 \\
Most Guessed (prior weighted) & 75\% & 92.1 \\
Most Common (`chance') & 38\% & 139.5 \\
\hline
\end{tabular}

\caption{For synthetic `bee' labelling data, we have data from the labelling of three non-experts and a ground truth expert. }
\label{cat_synth_table}
\end{table}

\begin{figure}[t!]
  \centering    
    \includegraphics[width=0.43\textwidth,trim=70 70 70 70,clip]{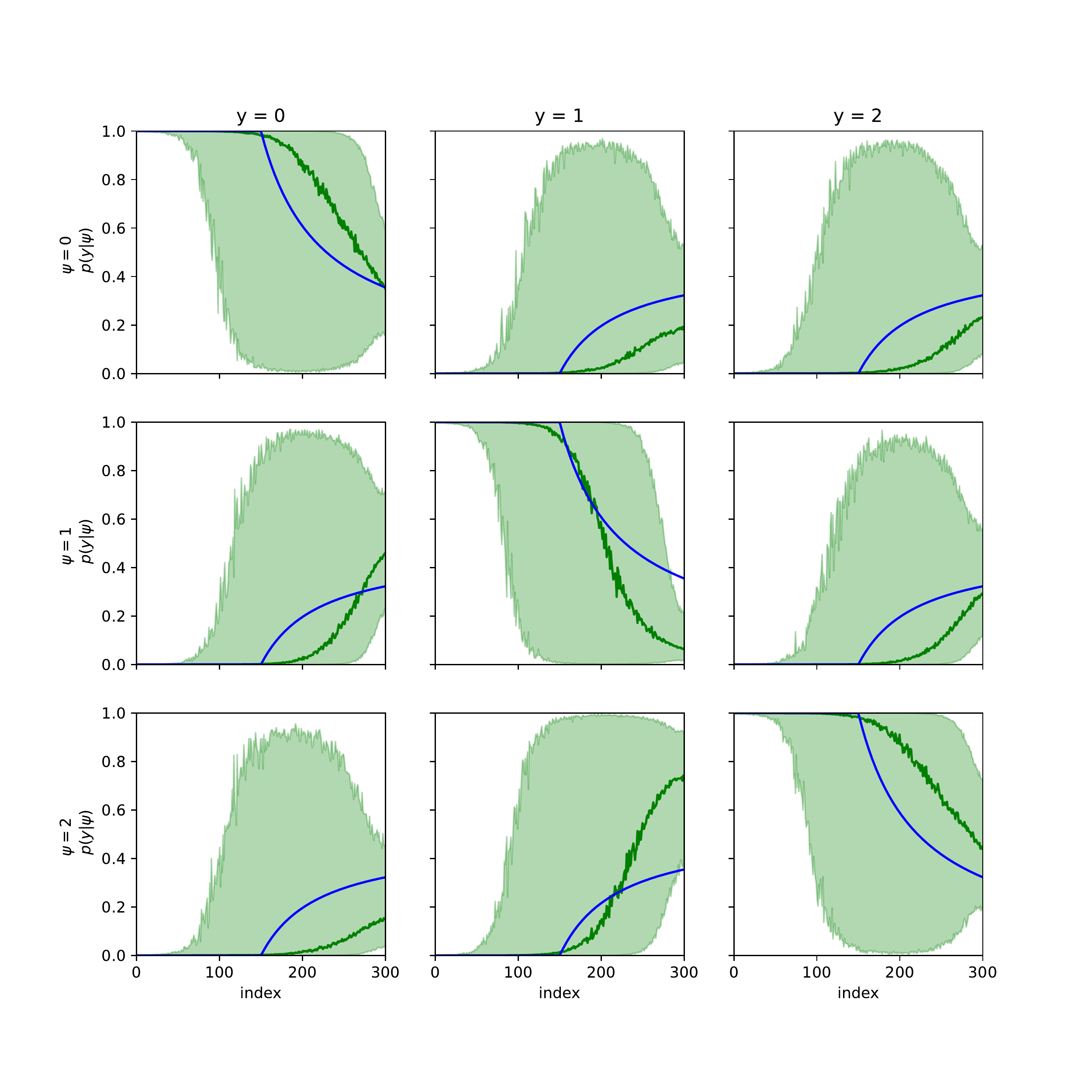}
  \caption{For the synthetic categorical dataset: The posteriors for the nine Gaussian processes that describe NE A's predictive ability: Blue line shows the synthetic generating function's `true' conditional probability, in green is the model's estimate with its 95\% CI. The probability refers to the likelihood of observing class $y$, given the true bee is of class $\psi$, with each row for a given \emph{true} bee ($\psi = 0,1,2$ respectively), while each column is the reported observed species. The x-axis in each plot indexes the 300 observations.}
    \vspace{-3mm}    
  \label{synth_person0}
\end{figure} 

\subsubsection{Bumblebee video classification}
\begin{figure}[t!]
  \centering
    \includegraphics[width=0.47\columnwidth]{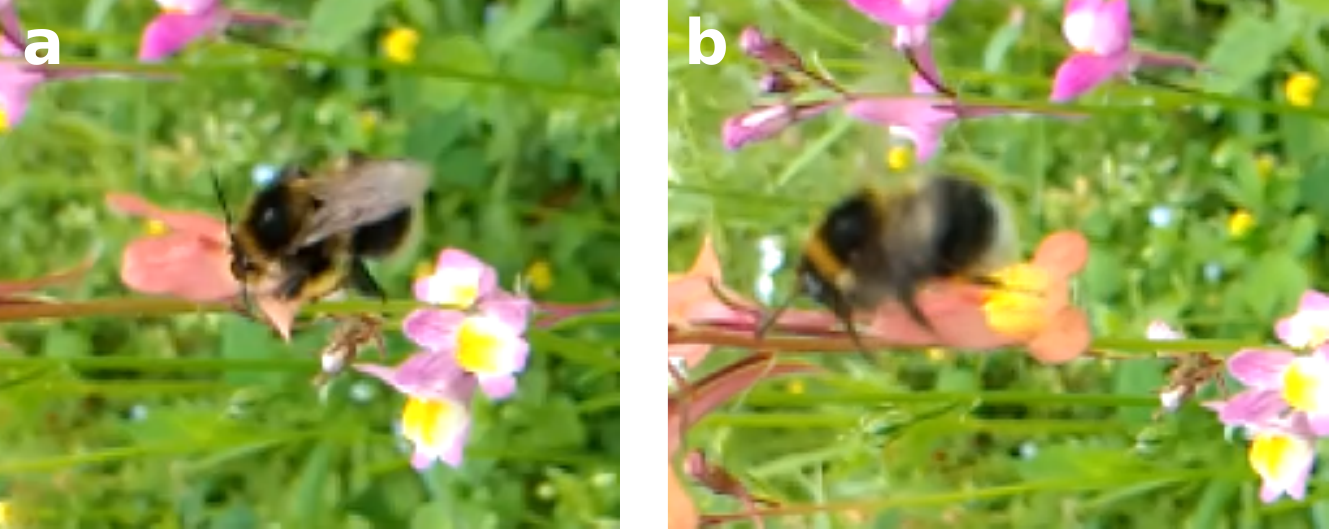}
  \caption{Parts of two frames from a video of a \textit{Bombus hortorum} foraging. In (a) the yellow bands on \emph{both} its thorax and abdomen are visible. In (b) one can also see the long face and tongue that distinguish it from \textit{Bombus jonellus}. It was sighted in northern England, making it unlikely to be \textit{Bombus ruderatus}. This video was labelled by two non-expert humans who identified it as (1) \textit{Bombus terrestris/lucorum}, (2) \textit{Bombus hortorum} and by the convolutional neural network as \textit{Bombus sylvestris}.}
    \vspace{-3mm}
  \label{beevideo}
\end{figure} 

We collected a dataset of 103 slow motion bee videos for a separate project \citep{ollett21}. These were labelled by an experienced and trained researcher `ground truth' and by six non-experts (NEs). These included five humans and one deep convolutional neural network (trained on photos from a public dataset). See \cite{ollett21} for details. \textit{Bombus terrestris} and \textit{Bombus lucorum} were combined into a single class as workers of these species are often visually indistinguishable. Those seven species which appear fewer than three times in the data were removed.\footnote{the list of removed bees: \textit{Bombus bohemicus}, \textit{Bombus jonellus}, \textit{Bombus sylvestris}, \textit{Bombus muscorum}, \textit{Bombus campestris}, \textit{Bombus monticola} and the non-bumblebee \textit{Apis mellifera}.} This included five species which only appeared in the NE labels. The result was 93 bees labelled by at least one NE (221 labels) and by ground truth. The NE accuracies and coverage varied considerably.\footnote{The five human NE accuracies (correct/total): 5/5, 56/60, 12/12, 21/25, 16/20 and the CNN NE: 62/89.}
The order the labelling was conducted was not known, so we decided to just use the Bias kernel.
We ran a 5-fold cross validation experiment. In spite of only using the Bias kernel we found the model still performed well. Table \ref{cat_bee_table} shows it achieves higher accuracy and with better uncertainty quantification than the other methods compared.

\begin{table}
\centering
\begin{tabular}{ c r r }
\hline
Method & Accuracy (\%) & NLPD \\
\hline
Calibration (Bias kernel) & 84\% & 59.9 \\
Collaborative Filtering & 58\% & inf \\
Most Guessed & 80\% & 69.4 \\
Most Guessed (`trust' weighted) & 82\% & 91.7 \\
Most Guessed (prior weighted) & 79\% & 98.4 \\
Most Common (`chance') & 54\% & 130.5 \\
\hline
\end{tabular}
\caption{Results for the real bee video labelling data. For context the CNN NE (the only NE to label all the videos) achieved 70\% accuracy.}
\label{cat_bee_table}
\end{table}

\section{Discussion}
In this paper we've proposed a method to calibrate a network of sensors using observations consisting of pairs of potentially bias or transformed observations in which a few observers are assumed to provide ground truth. We applied this to our main dataset - of static and mobile air pollution sensors in Kampala, adding a synthetic drift term to illustrate the potential issues in longer term data. We found the calibration pair method models the drift appropriately, making the predictions far more accurate. It also, importantly, provides uncertainty estimates on the calibration. For the real data without the synthetic drift, we found it had only a 1.5\% improvement in the MAE. Potentially over longer periods, we might expect the approach to become more relevant (as demonstrated by its utility when the drift was added). 
We also demonstrate how, replacing the likelihood function we can apply the method to a categorical dataset, in which the measurements are labels provided by citizen-scientists. We found the method was effective, and had good uncertainty quantification compared to other approaches.

Comparing the results of the multi-hop method and the variational calibration pair method, it seems that, for larger networks, the latter does better on both the synthetic and real data. It was somewhat surprising that it had much of an advantage in the Kampala case (especially with the synthetic drift added). In simulations we typically found the performance was considerably better than the multi-hop method when there were many sources of data to integrate. In the synthetic dataset for example, there were four reference sensors. So, with only one reference sensor, we anticipated the two approaches to perform similarly. However not only is the number of reference sensors relevant here, but also the number of paths between the reference sensor and the test sensor. Figure \ref{network1kampala}b indicates two routes exist from the reference sensor to the test sensor in the real dataset. Finally it isn't simply the path from the reference sensor to the test sensor that supports the variational result. The network graph doesn't show the full complexity of data availability over time, but importantly we envisage that the other sensors visited provide support. By also being visited, they can help constrain the calibration, indirectly. 
\subsection{Calibration in context}
Many studies exist in which low-cost air pollution sensors are co-located with reference sensors, and calibration functions are then derived to, in principle, allow the low-cost sensor to then be used without the reference sensor \citep{lewis2016evaluating, liu2017performance, zimmerman2018machine, barcelo2018calibrating, crilley2018evaluation, badura2019regression, wang2019calibration, datta2020statistical, lee2020long, ferrer2020multisensor}. This can also extend to calibrating remote observation data (for example \cite{shaddick2018data}). It has been noted many times that on-site field calibration is better than lab-based \citep{rai2017end}, and that sensor drift due to ageing or contamination means that regular recalibration is necessary \citep[][Section 3.2.3; and the results in this paper, Section \ref{kampaladata}]{crilley2018evaluation,marathe2021currentsense,castell2017can}. \cite{rai2017end} wrote,
\begin{quote}
...it is necessary to perform a field calibration for each sensor individually. Moreover, the calibration parameters might change over time depending on the meteorological conditions and the location, i.e., once the nodes are deployed it will be difficult to determine if they are under-or over-estimating the pollutant concentrations.
\end{quote}
Although it is widely agreed regular field recalibration is essential, the complexity of achieving this makes it potentially the main barrier to large-scale, low-cost sensor deployment \citep{rai2017end}. Several attempts have been made to look at how to do field-recalibration using mobile reference sensors \citep{hasenfratz2012fly,arfire2015model}. But using low-cost mobile sensors to act as `intermediaries' \citep{tsujita2004dynamic, hasenfratz2012fly} is far more practical. The low-cost sensors (e.g. the plantower pms5003) are typically smaller and often will be more resilient to being moved, and can be replaced more cheaply if damaged. In the AirQo Project in Kampala, moving the reference instruments regularly would have been very challenging. An alternative would be to swap the static low-cost sensors between reference colocation sites and their static locations. However such access required site permissions and expertise, while the mobile low-cost sensors could  visit the static stations simply by the motor-bike taxi drivers parking next to the sensor. This simplicity provided by permanently mobile low-cost sensors is discussed in \cite{zhao2021urban}. 

It is arguably only worth deploying, calibrating and maintaining the low-cost sensors if they can provide useful information to policy makers. \cite{castell2017can} suggest that they are limited to non-policy/science applications. They however looked at data from a city with relatively low-pollution. The average PM2.5 in Kampala was roughly 30-60 $\mu g/m^3$. The ten minute-averaged sample error in our experiment for the low cost sensor (with a calibration computed across the network) was about $15\mu g/m^3$. Once averaged to daily or annual estimates this would almost certainly be reduced below the 50\% threshold in the EU Data Quality Objectives (DQO) (for daily PM2.5 estimates). \citep{janssen2017guidance}. This is independent of the calibration approach and is probably possible in Kampala simply due to the high background pollution, reflecting the findings of \cite{castell2017can}, who found their relative accuracy improved at those times and locations with most pollution. This suggests achieving the EU's DQOs may be easier in high-pollution environments. It would be interesting to consider whether, in high-pollution locations, low-cost sensors could provide estimates of sufficient accuracy that they can be used for policy-making decisions. In-field calibration would still be necessary. By using a probabilistic approach to calibration and spatiotemporal modelling, one should also be able to state where (and when) in the domain the model reaches sufficient confidence.

A related field is the detection of sensor failure in a network of sensors. \cite{miskell2016data} used proxies (reference stations in an area with a similar land use) to detect potential sensor failure. Our method might be useful in the detection of drift, or similar signs of hardware failure. For example combining our method for calibration with a spatiotemporal model with leave one-sensor-out cross-validation. Poor prediction accuracy may be due to poor modelling assumptions or inference, e.g. we haven't included some aspect of the local environment; sensor failure, e.g. the inlet is clogged; or training data errors, e.g. if neighbouring sensors are malfunctioning. For more nuance it might be better to follow other approaches to fault detection. \cite{peng2017sensor}, for example, distinguish between types of fault.

Combining sensor estimates using a Bayesian approach is not novel. \cite{talampas2012maximum} take into account each sensor's uncertainty to combine their measurements using a Bayesian approach, and find it performs much better than simply averaging. In a different field \cite{perala2020calibrating} use Gaussian processes for another type of calibration problem: combining expert estimates, potentially with biases using a hierarchical GP. This could be applicable, in particular if different types of sensor can be grouped.

\subsection{The problem of spatially structured bias}
\cite{sousan2016evaluation} showed how different sources of particulate pollution lead to different responses by optical particle counters. 
\cite{castell2017can} and \cite{crilley2018evaluation} also discuss this issue. \cite{sandradewi2008using} propose that one deliberately use the change in the optical properties of the pollution to infer its source. We have assumed that 
the calibration function remains the same over space, and similar over short periods of time. If the former assumption doesn't hold we will need to colocate the reference instrument at each low-cost sensor location. If the latter assumption doesn't hold, then we can't use low-cost sensors at all.
\cite{chu2020spatial} suggest one could model the former, spatial-non stationarity. Presumably the temporal non-stationarity could be addressed in a similar way. They were investigating PM2.5 in a high humidity climate, so potentially applicable to our work in Kampala. Low cost OPC sensors are affected by high humidity, so the variation in humidity over Taiwan lead to changing biases. We could use these relevant features (such as humidity) in the calibration function. Unfortunately the \emph{type} of pollution (road dust, diesel exhaust or charcoal burning) isn't so easily measured, so some sort of spatio-temporal component might need to be introduced to the calibration function.

In the worst case we could imagine that one part of the city has pollution that is underestimated by all the network's low-cost sensors (e.g. a source such as burning tyres), thus all calibration that occurs between low-cost sensors in that area will fail to detect this bias. The only solution is to colocate a reference instruments in those locations. Carefully distributing the small number of reference instruments to sample a broad range of land-use types in the city is therefore prudent.

\subsection{Choice of model and choice of inference}
We didn't explore in this paper the potential capabilities of the joint model (in which the pollution is modelled explicitly in time and space), instead we focused on the `pair model' (in which the pollution remains a hidden, latent random variable) as we found it (in unreported work) more reliable than the joint model which had high correlations in the posterior mediated via both the prior \emph{and} each co-location (via the likelihood function). Conversely, the pair model scaled and behaved in a far more robust manner, with reliable optimisation (when using VI).

We have (in unreported work) experimented with using MCMC to approximate the variational distribution \citep{hensman2015mcmc} but found it failed to achieve timely mixing (probably due to the correlations in the posterior) when the network was more than two or three edges deep, so we concentrated on the VI approach.

The variational calibration pair model was quite successful in modelling the posterior variance. 
One reason might be that the approximating distribution included a full covariance matrix, a necessity due to the strong correlations between sensor parameters in the posterior. By avoiding the mean field approximation we might also have avoided some of the extreme underestimation of the variance \cite[see Figure 1 in][for an explanation]{blei2017variational}. We did (unreported) experiment with using the mean-field approximation in the joint model and found the variance often `collapsed', giving highly confident predictions, supporting this explanation.

One issue that might impact inference in larger networks is how the number of inducing points would roughly scale with $\text{time} (T) \times \text{sensors} (S) \times \text{parameters} (C)$, as we assume the number of inducing points is proportional to time. This would be even worse for the categorial example in which the number of parameters scales by the number of species-squared ($\numspecies^2$). This would pose both a computational challenge, with $O((C S T)^3)$ time complexity, but also leads to a huge number of parameters to express the variational distribution's covariance matrix, $\propto (CST)^2$. For the categorical model, we could alter our model to add an assumption of independence between species, making the inverse more tractable and reducing the number of parameters from $( \numspecies^2 S T)^2/2$ to $\numspecies (\numspecies S T)^2/2$.

Using fixed hyperparameters for our model was driven partly by domain knowledge that past stability in calibration does not guarantee future stability, but we also found them somewhat difficult to estimate using black box variational inference, an experience shared by others, e.g. \cite{nguyen2014automated} report that they `have found this to be problematic, ineffectual, and time-consuming.'

\subsection{Hetroscedastic likelihood}
In the likelihood used, the variance of the normal distribution was a fixed value. But some sensors (and locations) may have differing amounts of noise, and just as the calibration is expected to vary, so too will this likelihood variance. In the supplementary we explain how to include hetroscedastic noise \citep[see also, for example][]{lazaro2011variational}. We also propose that for the variational approximation of noise, one could use just a Dirac distribution (rather than a multivariate Gaussian) and so just specify the mean. The motivation being that estimating the uncertainty in the noise is unrealistic. Our implementation allows either a fixed likelihood variance, a  point estimate `Dirac' approximation, or a full multivariate Gaussian distribution, but in the examples in the paper we used the fixed version, as it seemed to perform sufficiently for our datasets.

\subsection{Summary and future work}
In summary, the calibration of a network through co-located observations of pairs of (potentially drifting) sensors can be performed by modelling each sensor's calibration function parameters using one or more Gaussian processes, and providing a likelihood function computed for \emph{two} observations. This provides a principled and robust Bayesian approach to handling this type of data. 
This will enable those designing sensor deployments to potentially use pair-wise calibration as part of the design solution, in which a reference sensor does not need to visit every low-cost sensor directly. 

To put this paper in context, \cite{cui2021new} describe a four stage calibration process for an air pollution monitoring system: (1 and 2) lab based, (3) field based reference sensor colocation (4) field based with occasional visits of mobile reference sensor. The calibration pair method would support the last of these four stages, allowing complex network structures, involving low-cost mobile sensors.

Future work includes optimising the choice of co-locations to minimise error: specifically choosing the best order to visit the static sensors in Kampala. There might be some sensors that are more useful to calibrate (due to their importance in the spatiotemporal model or location in the network). Cost is also a constraint: In our case, mobile-unit visits to the sensors had different (financial) costs, depending on their distance from the driver's base (known locally as their `stage'). In particular the reference sensors were quite distant.

The categorial bee-labelled problem has its own sampling-optimisation exploitation/exploration questions: deciding which videos to show which citizen-scientist/expert. We might wish to show a video to reduce the uncertainty in their calibration \emph{or} to show them bees that the model believes they can help classify \emph{or} maybe help improve their capabilities through training.

We do not consider uncertainty in the multi-hop model. We anticipate in future work looking at performing probabilistic inference over a simplified graph. We think this could be a good compromise (averaging over longer periods and making a relatively simplified graph network), potentially leading to faster inference than the variational Gaussian process approach, while still leading to good estimates of calibration and uncertainty, albeit in a more constrained modelling space.

The main limitation to future work in developing approaches towards networks of calibration the area of air pollution is that of data. Many low-cost network sensors exist \citep[e.g.][]{parmar2017iot,rai2017end,nyarku2018mobile,feinberg2019examining,zhao2021urban} but few perform regular co-location calibrations. Possibly this is because there are no well developed, easily used, computational tools to perform the in-field calibration discussed in this paper. Hopefully methods, such as the one outlined in this paper, will help solve this chicken-and-egg problem.

The example networks used in this paper (both synthetic and real) were not very deep. We could envisage that, with larger datasets (for example those collected on \url{www.zooniverse.org}) this network could become more complex, both potentially benefiting more from this method, but also providing a more challenging inference problem as the number of GPs grows.
One exciting possibility is to apply the method to the \url{beewalk.org.uk} project, a large, citizen-science initiative in which volunteers walk transects every month and count the number of each bumblebee species. If beewalkers were to occasionally walk other walkers' routes, could the data quality be improved through pair-wise calibration?

\subsection{Conclusion}
The problem of calibrating low-cost, large-scale, air pollution sensor networks is quickly becoming acute: A number of large scale deployments of low-cost air pollution networks have happened recently, including one running on 260 cars in Beijing \citep{zhao2021urban}. The difficulty in field-calibration and bounding the accuracy of estimates from these deployments suggests that methods that can provide automatic low-cost recalibration across the network are becoming increasingly necessary.

The calibration pair method described enables sensor deployments with complex networks of colocations. It allows the uncertainty in the predictions from these networks to be quantified, providing guaranteed confidence bounds and robust insights that mean the results from such networks can start being used by policy makers.

\section*{acknowledgements}
This work was part funded by google.org and an EPSRC GCRF grant, `Physically-informed probabilistic modelling of air pollution in Kampala using a low cost sensor network' [EP/T00343X/1].

\section*{conflict of interest}
The authors declare that they have no conflict of interest, and that there is no financial interest to report.

\section*{code and data}
The variational and multi-hop calibration methods are available as a python module which can be downloaded and installed from \url{https://github.com/lionfish0/calibration}, which includes a demonstration jupyter notebook, and the notebooks used to make the figures etc in this paper.

Kampala sensors: This data is owned and controlled by AirQo and is not available for public release. In particular, the raw boda boda data can reveal locations of individual journeys and the home locations of the taxi drivers. 

Bee videos and labels: we have uploaded these videos and labels to the University of Sheffield’s data repository (ORDA): \url{https://doi.org/10.15131/shef.data.19704538}.
\bibliography{refs}
\vfill
\pagebreak
\section*{Supplementary Material}
\begin{algorithm}[H]
\caption{Variational Inference for calibration. \\ Note: For implementation we structure an input matrix $\bm{X}$ to hold the time, sensor and component as three columns. This matrix is $2C$ times the length of $Y$. Split into $C$ submatrices, each pair of rows in each submatrix is associated with one row of $\bm{Y}$. $\bm{f}$ now becomes a vector with each item associated with one row of $\bm{X}$. The parameters are selected using slice notiation.}
\label{vialg}
 \hspace*{\algorithmicindent} {\textbf{Inputs}\\Observation time, sensor and parameter, $\bm{X}=\{\bm{x}_i\}_{i=1}^{2CN}$;\\Observation pair values, $\bm{Y} = \{[y_i^{(1)},y_i^{(2)}]\}_{i=1}^N$;\\Inducing point locations (time, sensor and parameter), $\bm{Z} = \{\bm{z}_i\}_{i=1}^M$;\\ Calibration function, $\phi(y,\bm{c})$;\\Number of parameters used by function, $C$;\\Kernel, $k(\cdot,\cdot)$;\\Likelihood variance $\sigma^2$;\\Reference sensors, $\bm{r}=\{0,1\}^S$;\\Number of samples in MC approximation, $P$.\\}
 \hspace*{\algorithmicindent}\textbf{Outputs}\\Approximating Gaussian distribution parameters: mean $\bm{m}$ and factor $\bm{R}$ (where distribution covariance is $RR^\top$).\\
\begin{algorithmic}[1]
\Procedure{VI}{$\bm{X}, \bm{Y}, \bm{Z}, \phi(\cdot), C, k(\cdot,\cdot), \sigma^2, \bm{r}$}
\While{$\bm{m}$ or $\bm{R}$ not converged}
\\ \Comment{$q(\bm{f})$ is the approximate posterior over all latent variables defined in $\bm{X}$.}
\State $q(\bm{f}) \sim N(K_{xz} K_{zz}^{-1} \bm{m}, K_{xx} - K_{xz}K_{zz}^{-1} K_{zx} + K_{xz}K_{zz}^{-1} RR^\top K_{zz}^{-1} K_{zx})$
\For {$j=1..P$} \Comment {Sample $P$ times}
  \For {$i=1..N$}
    \\ \Comment{Sample the latent variables relevant to the two sensors.}
    \State $\bm{s}^{(1)}_i,\bm{s}^{(2)}_i = \text{sample}[q(\bm{f}_{2i-1::2N}), q(\bm{f}_{2i::2N})]$  \Comment{$\bm{s}_i^{(1)}$ and $\bm{s}_i^{(2)}$ are each of length $C$.}
    \State $L_{ij} = \log p\Bigg(\begin{bmatrix}y_{i}^{(1)}\\y_{i}^{(2)}\end{bmatrix} \Bigg| \begin{bmatrix}\bm{s}_i^{(1)} \\ \bm{s}_i^{(2)}\end{bmatrix}\Bigg)$ \Comment{Compute likelihood of sample, using \eqref{likelihood}} 
  \EndFor
\EndFor
\State $\mathcal{L} \gets \frac{1}{P} \sum_{j=1}^P \sum_{i=1}^N \left(L_{ij}\right) - D_{KL}\Big(N(\bm{m},RR^\top),N(0,K_{ZZ})\Big)$ \Comment{Compute ELBO}
\State compute $\frac{d\mathcal{L}}{\bm{m}}$ and $\frac{d\mathcal{L}}{\bm{R}}$  \Comment{using automatic differentiation.}
\State update $\bm{m}$ and $\bm{R}$ using stochastic gradient descent (using Adam).
\EndWhile
\EndProcedure
\end{algorithmic}

\end{algorithm}

\begin{algorithm}
\caption{Prediction for Algorithm \ref{vialg}}
 \hspace*{\algorithmicindent} {\textbf{Inputs}\\Test time, sensor and component, $\bm{X}_*=\{\bm{x}_{*i}\}_{i=1}^{N_*}$; $\bm{y}_*=\{{y}_{*i}\}_{i=1}^{N_*}$ raw observations (unlike normal regression we need to give uncalibrated observations at test time) \\Inducing point locations (time, sensor and component), $\bm{Z} = \{\bm{z}_i\}_{i=1}^M$;\\ Calibration function, $\phi(y,\bm{c})$;\\Number of components used by function, $C$;\\Kernel, $k(\cdot,\cdot)$;\\Approximating Gaussian distribution parameters: $\bm{m}$ and $\bm{\bm{R}}$.\\Number of samples for each test point, $P$.\\}
 \hspace*{\algorithmicindent}\textbf{Outputs}\\An $N_* \times P$ matrix of $P$ samples for each of the $N_*$ test points, $\bm{S}$\\
\begin{algorithmic}[1]
\Procedure{Predict}{$\bm{X}_*, \bm{Y}_*, \bm{Z}, \phi(\cdot), C, k(\cdot,\cdot)$}
\\ \Comment{$q(\bm{f})$ is the approximate posterior over all latent variables defined in $\bm{X}_*$.}
\State $q(\bm{f}) \sim N(K_{x_*z} K_{zz}^{-1} \bm{m}, K_{x_*x_*}-K_{xz}K_{zz}^{-1} K_{zx_*}+K_{x_*z}K_{zz}^{-1} RR^\top K_{zz}^{-1} K_{zx_*})$
\For {$i=1..N_*$}
    \\ \Comment{Sample the latent variables relevant to each test point.}
    \State $\bm{s}_i = \phi(y_i,\text{sample}[q(\bm{f}_{i::N_*})])$ \Comment Sample $K$ times.
\EndFor
\EndProcedure
\end{algorithmic}
\label{alg1pred}
\end{algorithm}

\begin{algorithm}
\caption{Multi-hop `Graph' algorithm}
\label{mhga}
 \hspace*{\algorithmicindent} {\textbf{Inputs}\\Observation time, sensor id pair, $\bm{X}=\{\bm{x}_i\}_{i=1}^N$, so $\bm{X}$ is $(N \times 3)$;\\Observation pair values, $\bm{Y} = \{[y_{i1},y_{i2}]\}_{i=1}^N$, so $\bm{Y}$ is $(N \times 2)$;\\Window size, $\delta$;\\Edge `distances', connect by colocation events and over time, $d_\text{colocation}$, $d_\text{time}$;\\Reference sensors, $\bm{r}=\{0,1\}^{|\numspecies|}$.\\}
 \hspace*{\algorithmicindent}\textbf{Outputs}\\Dictionary of scaling factors for (sensor,window) tuples $\bm{F}$.\\ 
\begin{algorithmic}[1]
\Procedure{BuildGraph}{$\bm{X}, \bm{Y}, \delta, \bm{r}$}
\For {$w$ in all time windows $W$ of width $\delta$}
  \For {$s_i$,$s_j$ in all pairs of sensors from $X$, where $s_i \neq s_j$}
    \State $Y' \gets Y_{X_{:,2}=s_i \land X_{:,3}=s_j \land X_{:,1} \in w}$ \Comment{Selects observation pairs that are in the time window and between sensors $s_i$ and $s_j$}
    \If {$|Y'| \geq 5$} \Comment{We don't add edges where there are fewer than five observations}
      \State G.addEdge($(s_i,w)\to(s_j,w)$, value=mean($\log(Y'_{:,1})-log(Y'_{:,2})$, distance=$d_{colocation}$))
    \EndIf      
  \EndFor
\EndFor
\For {$w$ in all time windows $W-1$ of width $\delta$}
  \For {$s_i$ in all sensors from $X$}
    \State G.addEdge($(s_i,w)\to(s_i,w+1)$, value=0, distance=$d_{time}$)
    \State G.addEdge($(s_i,w+1)\to(s_i,w)$, value=0, distance=$d_{time}$)
  \EndFor
\EndFor
\For {$w$ in all time windows $W-1$ of width $\delta$}
  \For {$s_i$ in all sensors from $X$}
    \State $P \gets \text{Shortest path (using Dijkstra) from}\; (s_i,w)\; \text{to any reference sensor, specified in}\; r.$
    \If {Path $P$ exists}
      \State $F[s_i,w] = \sum_{p \in P}{p_{value}}$ \Comment{Sum log ratios over path}
     \EndIf
  \EndFor
\EndFor    
\EndProcedure
\end{algorithmic}
\end{algorithm}

\begin{algorithm}
\caption{Multi-hop `Graph' prediction algorithm}
\label{mhgpa}
 \hspace*{\algorithmicindent} {\textbf{Inputs}\\Observation time, sensor id, $\bm{X}_*=\{\bm{x}_i\}_{i=1}^N$, so $\bm{X}_*$ is $(N_* \times 2)$;\\Observed raw values, $\bm{y}_{\text{raw}}$, so $\bm{y}_{\text{raw}}$ is an $(N_* \times 1)$ vector;\\Window size, $\delta$;\\Dictionary of scaling factors for (sensor,window) tuples $\bm{F}$, from Algorithm \ref{mhga}.\\}
 \hspace*{\algorithmicindent}\textbf{Outputs}\\Predicted calibrated values, $\bm{y}_{*}$, so $\bm{y}_{*}$ is an $(N_* \times 1)$ vector.
\begin{algorithmic}[1]
\Procedure{Predict}{$\bm{X}_*, \bm{y}_{\text{raw}}, \delta, \bm{r}$, $G$}
\For {$\bm{x}_*$ in $\bm{X}_*$; $y_*$ in $\bm{y}_*$; and $y_\text{raw}$ in $\bm{y}_\text{raw}$}
  \State $w \gets \text{window computed for} [\bm{x}_*]_1 \text{using window size}\; \delta$
  \If{$([\bm{x}_*]_2,w) \notin F$}
  \State Raise exception: No path to a reference sensor.
  \EndIf
  \State $y_* \gets e^{F[[\bm{x}_*]_2,w]} \times y_\text{raw}$
  
\EndFor
\EndProcedure
\end{algorithmic}
\end{algorithm}
\end{document}